\documentclass[sigconf,10pt]{acmart}
\setcopyright{rightsretained}

\copyrightyear{2021}
\acmYear{2021}
\setcopyright{acmlicensed}\acmConference[SoCC '21]{ACM Symposium on Cloud Computing}{November 1--4, 2021}{Seattle, WA, USA}
\acmBooktitle{ACM Symposium on Cloud Computing (SoCC '21), November 1--4, 2021, Seattle, WA, USA}
\acmPrice{15.00}
\acmDOI{10.1145/3472883.3487001}
\acmISBN{978-1-4503-8638-8/21/11}

\usepackage[makeroom]{cancel}
\usepackage{ifthen}
\usepackage{xspace}
\usepackage{enumitem}
\usepackage{graphicx}
\usepackage{subcaption}
\usepackage{amsmath}
\usepackage{dsfont}
\usepackage{footmisc,longtable}
\usepackage[compact,small]{titlesec}
\usepackage{etoolbox}
\usepackage{mathbbol}
\usepackage{bbm}

\usepackage{csquotes}

\usepackage{array}
\usepackage{pifont}
\usepackage{marvosym}

\graphicspath{{figs/}{data/}}

\newcommand{\anonymize}{false}
\newcommand{\showComments}{false}
\newcommand{\showCommentsNew}{true}

\newcommand{\addcoreapp}{0} 

\newboolean{comments}
\setboolean{comments}{\showComments}
\newboolean{commentsNew}
\setboolean{commentsNew}{\showCommentsNew}

\newboolean{anon}
\setboolean{anon}{\anonymize}
\newcounter{coreapp}
\newcommand{\sysname}{\textsc{Sayer}\xspace}
\ifthenelse{\boolean{anon}} {
\newcommand{\azure}{Cloud\xspace}
} {
\newcommand{\azure}{Azure\xspace}
}
\newcommand{\health}{\azure-Health\xspace}
\newcommand{\scale}{\azure-Scale\xspace}

\newcommand{\sone}{\textit{S1}\xspace}
\newcommand{\stwo}{\textit{S2}\xspace}
\newcommand{\sthree}{\textit{S3}\xspace}

\newcommand{\ie}{i.e.,\xspace}
\newcommand{\eg}{e.g.,\xspace}

\newcommand{\eps}{\ensuremath{\epsilon}\xspace}

\newcommand{\Exp}{\mathbb{E}}
\newcommand{\Var}{\mathbb{V}}
\newcommand{\algofont}[1]{\ensuremath{\mathtt{#1}}}
\newcommand{\epsgreedy}{\algofont{EpsilonGreedy}\xspace}
\newcommand{\ips}{\algofont{IPS}\xspace}

\newcommand{\implicit}{\algofont{Implicit}\xspace}

\newcommand{\tuplerl}{\ensuremath{(\vec{x},a,c,p)}\xspace}
\newcommand{\tupleim}{\ensuremath{(\vec{x},a,c,\vec{p})}\xspace}
\newcommand{\vw}{\algofont{VW}\xspace}


\newcommand{\mypara}[1]{\vspace{0.1cm}\noindent{\bf {#1}.}~}
\newcommand{\myparaitalic}[1]{\vspace{0.07cm}\noindent{\em {#1}}~}

\def\1{\mathbb{1}}
\def\Exp{\mathbb{E}}

\setlist{leftmargin=*,itemsep=0pt,parsep=0pt,topsep=1pt,partopsep=1pt}

\ifthenelse{\boolean{comments}} {
  \newcommand{\sid}[1]{\textcolor{blue}{(Sid: #1)}}
  \newcommand{\alex}[1]{\textcolor{magenta}{(Alex: #1)}}
  \newcommand{\mathias}[1]{\textcolor{olive}{(Mathias: #1)}}
  \newcommand{\amit}[1]{\textcolor{green}{(Amit: #1)}}
  \newcommand{\junchen}[1]{\textcolor{orange}{(Junchen: #1)}}
  \newcommand{\mn}[1]{\textcolor{violet}{(Mihir: #1)}}
  \newcommand{\sh}[1]{\textcolor{orange}{(Sang Hoon: #1)}}
} {
  \newcommand{\sid}[1]{}
  \newcommand{\alex}[1]{}
  \newcommand{\mathias}[1]{}
  \newcommand{\amit}[1]{}
  \newcommand{\junchen}[1]{}
  \newcommand{\mn}[1]{}
  \newcommand{\sh}[1]{}
}
\ifthenelse{\boolean{commentsNew}} {
  \newcommand{\sidn}[1]{\textcolor{blue}{(Sid: #1)}}
  \newcommand{\mathiasn}[1]{\textcolor{olive}{(Mathias: #1)}}
  \newcommand{\junchenn}[1]{\textcolor{orange}{(Junchen: #1)}}
  \newcommand{\shn}[1]{\textcolor{orange}{(Sang Hoon: #1)}}
} {
  \newcommand{\sidn}[1]{}
  \newcommand{\mathiasn}[1]{}
  \newcommand{\junchenn}[1]{}
  \newcommand{\shn}[1]{}
}

\newcommand{\ignore}[1]{}

\newcounter{packednmbr}

\newenvironment{packeditemize}{\begin{list}{$\bullet$}{\setlength{\itemsep}{0.5pt}\addtolength{\labelwidth}{-4pt}\setlength{\leftmargin}{2ex}\setlength{\listparindent}{\parindent}\setlength{\parsep}{1pt}\setlength{\topsep}{2pt}}}{\end{list}}

\newcounter{insightlabel}
\newcounter{insightnmbr}
\renewcommand{\theinsightlabel}{\textbf{\theinsightnmbr}}

\newcommand{\tightcaption}[1]{\vspace{-0.1cm}\caption{#1}\vspace{-0.25cm}}

\begin{document}

\date{}

\title[\sysname: Using Implicit Feedback to Optimize System Policies]{\huge \sysname: Using Implicit Feedback to Optimize System Policies}

\author{
  {\rm Mathias L\'ecuyer$^{\star4}$}, {\rm Sang Hoon Kim$^{*1}$}, {\rm Mihir Nanavati$^{\star3}$}, {\rm Junchen Jiang$^5$},\\{\rm Siddhartha Sen$^2$}, {\rm Aleksandrs Slivkins$^2$}, {\rm Amit Sharma$^2$}\\
  \vspace{0.15cm}
  {\large
   {\em $^1$Columbia University}, {\em $^2$Microsoft Research}, {\em $^3$Twitter}, {\em $^4$University of British Columbia}, {\em $^5$University of Chicago}
   }
} 
\thanks{$^{\star*}$Current affiliations. Work done previously, at: $^\star$Microsoft Research, and $^*$University of Chicago.}

\renewcommand{\shortauthors}{
M. L\'ecuyer, S.H. Kim, M. Nanavati, J. Jiang, S. Sen, A. Slivkins, A. Sharma
}

\settopmatter{printfolios=true}
\begin{abstract}

We observe that many system policies that make threshold decisions involving a 
resource (\eg time, memory, cores) naturally reveal additional, or {\em implicit}
feedback. For example, if a system waits $X$ min for an event to occur, then it
automatically learns what would have happened if it waited $< X$ min, because 
time has a cumulative property. This feedback
tells us about alternative decisions, and can be used to improve the
system policy. However, leveraging implicit feedback is difficult because it
tends to be one-sided or incomplete, and may depend on the outcome of the event.
As a result, existing practices for using feedback, such as simply
incorporating it into a data-driven model, suffer from bias.

We develop a methodology, called \sysname, that leverages implicit feedback to evaluate 
and train new system policies. \sysname builds on two ideas from reinforcement
learning---randomized exploration and unbiased counterfactual \newline 
estimators---to leverage data
collected by an existing policy to estimate the performance of new candidate policies,
without actually deploying those policies.
\sysname uses {\em implicit exploration} and {\em implicit data augmentation}
to generate implicit feedback in an unbiased form, which is then used by
an {\em implicit counterfactual estimator} to evaluate and train new policies. 
The key idea underlying these techniques is to assign implicit probabilities to decisions 
that are not actually taken but whose feedback can be inferred; these probabilities are carefully 
calculated to ensure statistical unbiasedness. 
We apply \sysname to two production scenarios in \azure, and show that it can evaluate
arbitrary policies accurately, and train new policies that outperform the
production policies.

\end{abstract}

\maketitle
\pagenumbering{gobble}


\section{Introduction}
\label{sec:intro}

A {\em system policy} is any logic that makes decisions for a system, such as choosing a
configuration, setting a timeout value, or deciding how to handle a request.
System policies are pervasive in cloud infrastructure and can seriously impact the
performance of a system. For this reason, developers are constantly trying to iterate on
and improve them. Typically this is done by collecting feedback from the
currently deployed policy and analyzing or processing it to generate new candidate
policies. The more feedback that can be collected, the more insights that can be learned
to improve the policy.

We observe that a large class of system policies naturally reveal additional feedback
beyond the decision that is actually made. These policies make {\em threshold decisions}
involving a resource such as time, memory, or cores. Since the resources are naturally
cumulative, thresholding on one value often reveals feedback for other values as well.
For example, consider a system policy that decides how long to wait for an unresponsive machine
before rebooting it. If the policy waits $X$ min, then it automatically learns what would
have happened if it waited $<$$X$ minutes, because time is cumulative. We call this kind of 
feedback {\em implicit feedback}.

Ideally, we would like to leverage implicit feedback to evaluate new candidate policies
and ask:
what would happen if we deployed this policy in production? This type of ``what-if''
question can be answered using {\em counterfactual evaluation}, a process that uses
data collected from a deployed policy to estimate the performance of a candidate policy. If
counterfactual evaluation can be done offline, it provides a powerful alternative to
methods like A/B testing (which require live deployment of a policy), because it means that
we can evaluate policies without ever deploying them~\cite{horvitz1952generalization}.

Unfortunately, implicit feedback is difficult to leverage 
because it typically appears in a biased form. For instance
in the above example, feedback is only received for wait times $<$$X$, \ie it is one-sided. 
In fact, the amount of feedback received may even depend on the outcome of an event: if the 
unresponsive machine recovers within $X$ min, then we actually get feedback for any wait time, 
because we know exactly when the machine recovered.
Biased feedback is difficult to leverage for counterfactual evaluation because it
generates more feedback for some actions than for others. For example, a policy that
always waits $\le$3 min will never generate feedback that can be used to evaluate a 
candidate policy that waits 4 min or more.

How do we leverage implicit feedback that is one-sided and outcome-dependent?
We draw on ideas from statistics and reinforcement learning (RL) to
provide a starting point.
Specifically, we leverage: {\em randomized exploration}, which modifies a
deployed policy to choose a random action some of the time, thereby increasing
the coverage over all actions; and {\em unbiased counterfactual estimators},
which use exploration data collected from a policy to accurately estimate a
candidate policy's performance.
The threshold decisions we study naturally satisfy certain independence
assumptions (\S\ref{subsec:design:overview}), allowing us to build on
particularly efficient exploration algorithms and counterfactual estimators.
Unfortunately, all of these techniques assume that a policy receives feedback
only for the (single) action that it takes. In other words, they are unable to
leverage implicit feedback.

We develop a methodology called \sysname that leverages implicit feedback to perform
unbiased counterfactual evaluation and training of system policies.
\sysname develops three techniques to harness the implicit feedback revealed by threshold
decisions:
(1) {\em implicit exploration} builds on existing exploration algorithms to maximize the
amount of implicit feedback generated; (2) {\em implicit data augmentation} augments the
logged data from a deployed policy to include implicit feedback; and (3) an {\em implicit
counterfactual estimator} uses the augmented data to evaluate and train new policies in an
unbiased manner. A key idea underlying \sysname is to assign implicit probabilities to
decisions that are not actually taken but whose feedback can be inferred; these
probabilities are carefully calculated to ensure statistical unbiasedness.

As a community, we have developed a variety of methods for counterfactual evaluation,
ranging from offline methods like simulators and data-driven models, to online methods
like A/B testing. However, as we discuss in \S\ref{subsec:counterfactual}, most of these
approaches suffer from bias, and approaches that are unbiased are either too invasive
(\eg they require a live deployment) or are not data-efficient. 
None of these approaches leverage implicit feedback. 
In contrast, \sysname builds on
techniques from RL to enable unbiased and data-efficient
counterfactual evaluation.
 
\mypara{Contributions} \sysname makes the following contributions:

\begin{itemize}

    \item We demonstrate the presence of implicit feedback in
    system policies that make threshold decisions (\S\ref{sec:motivation}), and develop
    a framework for harnessing this feedback. \sysname
    provides a new unbiased counterfactual estimator for
    implicit feedback, supported by new implicit feedback-aware
    algorithms for exploration and data augmentation (\S\ref{sec:design}).
    
    \item We develop an architecture for integrating \sysname into the 
    lifecycle of existing system policies (\S\ref{subsec:design:overview}),
    supporting their continuous optimization lifecycle (\S\ref{subsec:design:components}).
    This enables policies to continuously evolve. 

    \item We apply \sysname to two production scenarios in \azure:
        (\S\ref{sec:health}-\S\ref{sec:scale}):
        \health, a system that handles unresponsive machines, and
        \scale, a system that creates scalable sets of virtual
        machines.
\end{itemize}


\section{Implicit Feedback}
\label{sec:motivation}

This section provides the necessary background for studying system policies that yield
implicit feedback.
We first show that threshold decisions naturally reveal additional, implicit feedback
beyond the decision that is actually taken (\S\ref{subsec:implicit}).
Leveraging this feedback to evaluate new policies is difficult, however,
because it appears in a biased form (\S\ref{subsec:challenges}).
A survey of existing approaches for policy evaluation (\S\ref{subsec:counterfactual})
shows that reinforcement learning (RL) provides a foundation for addressing this bias.
However, no existing approach, including RL ones, can leverage implicit feedback,
motivating the goals of \sysname (\S\ref{subsec:goals}).

\subsection{Implicit Feedback in Threshold Decisions}
\label{subsec:implicit}

Many system policies make {\em threshold decisions} based on a resource such as time,
memory, cores, etc.. These decisions choose a threshold value of the resource on which some
behavior of the system is conditioned. Some real examples of threshold decisions made in the \azure cloud include:

\begin{packeditemize}
  \item {\em time:} How long to wait for unresponsive machines before rebooting them (\health, \S\ref{sec:health}).
  \item {\em VMs}: How many extra VMs to create in order to complete a set of VM creations faster (\scale, \S\ref{sec:scale}).
  \item {\em cores: } What level of CPU utilization triggers an elastic scaler to increase/decrease the number of replicas?
\end{packeditemize}

Decisions like these are pervasive in cloud infrastructure and can seriously impact the
performance of a system. We study two of these decisions in this paper.
As a running example we use \health, a system which monitors the
health of physical machines in \azure's datacenters, with the goal of minimizing downtime
for customer VMs. If a machine becomes unresponsive, a system policy decides how
long to wait for the machine to recover before rebooting it and reprovisioning its VMs, a process
that may take $>$10 min. The policy chooses a wait time from $\{1,2,\ldots,10\}$ min (the
longest wait time being comparable to the reboot cost). 

In each of the above examples, the resource in question has a natural cumulative property.
For example, if the \health policy waits $X$ min for an unresponsive machine, this
automatically tells us what would have happened if it waited $<$$X$ min, because time
is cumulative. Similarly, allocating $X$ extra VMs tells 
us what would have happened if we allocated fewer VMs,
and scaling up at $X\%$ CPU utilization tell us about 
lower thresholds.
Interestingly, feedback can also be revealed for thresholds greater than the chosen action.
For example in \health, if we wait $X$ min for an unresponsive machine and it
recovers within that time, then we actually get feedback for all other waiting
times, since we know exactly when the machine recovers. In this case, the
amount of feedback depends on the outcome of an event.

In all of the above examples, feedback is revealed for a decision that was not actually taken,
but can be inferred from the decision that was taken. We call this {\em
implicit feedback}.


\subsection{The Bias Problem}
\label{subsec:challenges}

Implicit feedback has the potential to accelerate improvements to system
policies, because it allows
developers to ask ``what if'' questions about alternative policy decisions.
For example, an \health developer may ask: what if we waited longer for all
unresponsive machines, or what if we waited less time for newer machines than older
ones?
Answering these questions requires {\em counterfactual evaluation}, which evaluates the performance of a new {\em
candidate policy} using feedback collected from the current {\em deployed policy}.

\begin{figure}[t]
  \centering
	\includegraphics[width=0.8\linewidth]{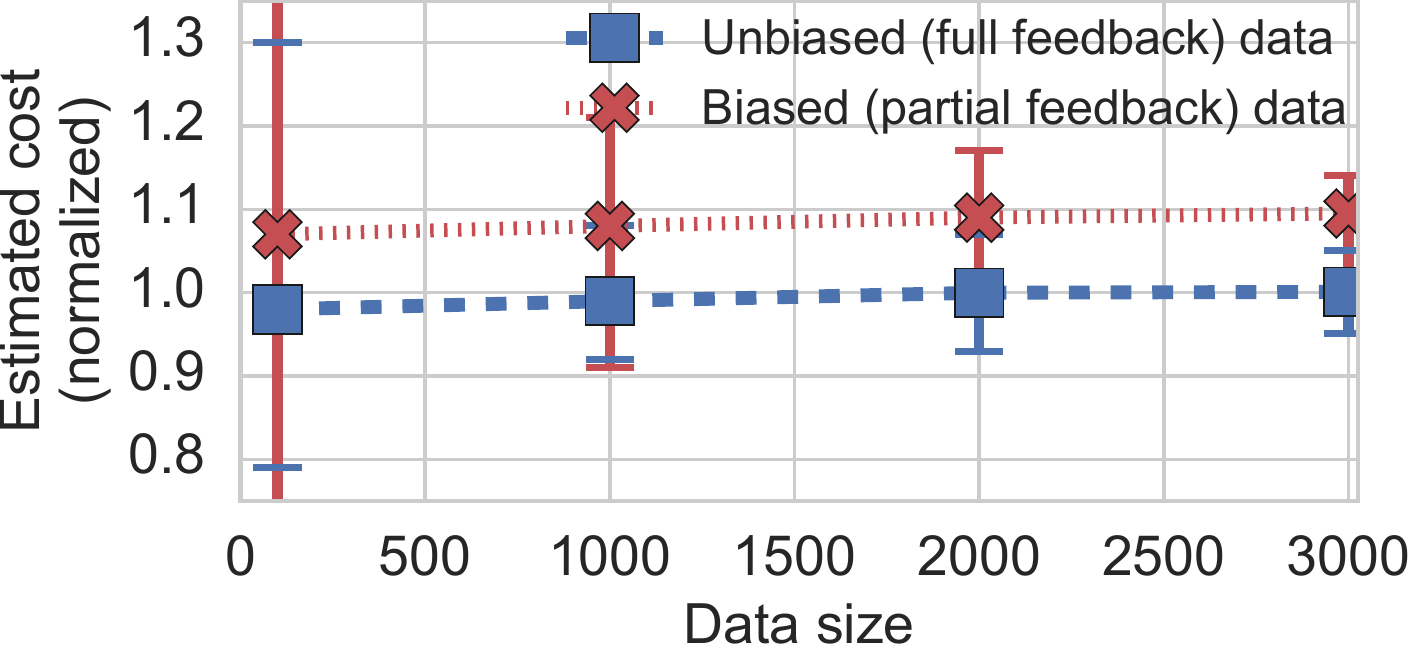}
  \tightcaption{Counterfactual evaluation of a candidate policy using unbiased vs. biased
  data, based on real production data from \health.
  The mean estimates are normalized to the candidate
  policy's true cost. Estimates based on biased data misrepresent the true cost by
  9\%, while estimates based on unbiased data are accurate and have lower variance.
  Confidence intervals are obtained by repeating the analysis on subsamples of the same
  size using a bootstrap procedure~\cite{efron1979bootstrap}.}
  \label{fig:bias-prod}
\end{figure}

\mypara{Partial feedback}
Unfortunately, implicit feedback is often one-sided---\eg feedback is revealed
for all thresholds $\le$$X$ but not $>$$X$---making it a form of {\em partial
feedback}, where each decision only reveals feedback for some of the
actions.
Partial feedback is inherently {\em biased}, because it
generates more feedback for some actions than others, making it difficult to use
for counterfactual evaluation.
For example, an \health policy that always waits $\le$$3$ min for unresponsive
machines will never generate feedback for longer waiting times. As a result, any
data collected from this policy cannot be used to counterfactually evaluate a
candidate policy that waits 4 min or more.

\mypara{Full feedback}
In contrast, {\em full feedback} arises when each decision reveals feedback for
every possible action.
This would be the case if the \health policy waits the maximum time (10 min) for
every machine, revealing implicit feedback for all lower times as well. Such full
feedback data is {\em unbiased}.
Unfortunately, full feedback data is rare in threshold decisions, and typically
only arises when a system is initially deployed with a very conservative policy
(often to collect data that will be used to train better policies).
In our work, we use full feedback data only to train initial policies, and as an
idealized baseline to quantify the effects of biased data.

As an illustration of the bias problem, we take production data from an initial two-month period during which
\health deployed the conservative policy of always waiting 10 min, generating an unbiased, full
feedback dataset.
We then derive a biased, partial feedback dataset based on a more optimized policy they used later on.
We use the biased and unbiased datasets to counterfactually evaluate a new candidate policy.
Figure~\ref{fig:bias-prod} shows that the counterfactual estimate 
using unbiased data closely matches the true cost of the candidate
policy, even when the dataset is small;
with more data, the variance reduces sharply around the true cost. In contrast, when using biased data, the estimated cost
deviates from the true cost by 9\%, and no amount of additional data removes this bias.
This is an important point: {\em although additional data can reduce the variance of an
estimate, it cannot remove the underlying bias}~\cite{Langford-scavenge}.
The candidate policy in this example appears to have higher cost than
it truly does, which could mislead the \health team.

\subsection{Existing Approaches}
\label{subsec:counterfactual}

Many existing approaches aim to address the bias problem when counterfactually 
evaluating a policy, as summarized in Table~\ref{tbl:approaches}.
Here, we discuss the strengths/weaknesses of each approach 
in the context of threshold decisions and their implicit feedback.
The high-level takeaway is that most approaches used in systems still suffer from bias.
Unbiased approaches from RL give a strong starting point for \sysname, but they are
are either too invasive (\eg they require a live deployment) or are not
data-efficient.
In particular, no existing approach leverages implicit feedback.

\begin{table}[t] \footnotesize
\begin{tabular}{r >{\centering\arraybackslash}m{1.1cm} >{\centering\arraybackslash}m{1.5cm}>{\centering\arraybackslash}m{1.8cm}
>{\centering\arraybackslash}m{1.8cm}}
\hline
 & Low bias & Data efficient & Less invasive \\\hline\hline
A/B testing & \ding{52} & \ding{56} & \ding{56} \\\hline
Online learning & \Circpipe & \ding{56} & \ding{56} \\\hline
Simulator  & \ding{56} & \ding{52} & \ding{52} \\\hline
Data-driven modeling & \ding{56}  & \ding{52} & \ding{52} \\\hline
Naive exploration/IPS & \ding{52} & \ding{56} & \ding{52} \\\hline
{\bf \sysname} & \ding{52} & \ding{52} & \ding{52} \\\hline \\
\end{tabular}
\caption{Different approaches for counterfactually evaluating policies that make threshold decisions (\Circpipe~$=$~``somewhat'').
Naive exploration and \ips allow unbiased, offline evaluation of arbitrary policies, 
but they fail to leverage 
implicit feedback like \sysname. 
\vspace{-1.0cm}}
\label{tbl:approaches}
\end{table}

{\bf A/B testing} is the gold standard for evaluating policies in a cloud
system~\cite{KohaviLSH09,KohaviAB-2015}, but it requires deploying each candidate policy live
alongside the current deployed policy, and randomly splitting traffic/requests between the
policies. The data collected from an A/B test can only be used to evaluate the deployed
policies, making it a costly and inefficient approach. {\bf Online learning} approaches
also deploy a policy in a live setting and use data collected from its decisions to
continuously update the policy.
Several systems~\cite{dong2015pcc, dong2018pcc, tesauro2007reinforcement,
mao2017neural, erickson2010effective, alipourfard2017cherrypick} 
use online learning or RL algorithms that can accurately evaluate
candidate policies that are very similar to the depoyed policy, but incur
bias when evaluating policies that are different.

Instead of deploying policies, a {\bf simulator} of the production environment can be used to
evaluate arbitrary policies offline. This approach is
data-efficient and non-invasive, but creating and maintaining an accurate simulator of
a complex, evolving system can be as large an undertaking as the system
itself~\cite{floyd2001difficulties, bartulovic2017biases}, making it prone to modeling
biases~\cite{floyd2001difficulties}.

Because of these difficulties, system designers often create {\bf
data-driven models} to predict the outcome of a given decision in a live system
(\eg~\cite{jiang2016webperf,tariq2008answering,krishnan2013video}).
In particular, many proposals train ML models to predict the outcome
of an action based on all the available context, and make decisions based on these
predictions~\cite{zhu2017bestconfig,fu2021use,van2017automatic,li2019qtune,zhang2019end, bingham2018pyro,klimovic2018selecta,peng2018optimus,venkataraman2016ernest,yadwadkar2017selecting}.
However, a misspecified model will introduce biases, and (re)training it on partial
feedback data collected when following the decisions of an earlier model will 
simply perpetuate these biases (see Figure \ref{fig:bias-prod}).

Another data-driven modeling approach, currently used by the \health team, is survival
analysis.
Survival analysis is a statistical modeling approach that postulates a
distribution over machine recovery times and uses this to predict any unobserved
outcomes.
This distribution is fitted to the data using right censoring, which explicitly
models the probability of missing feedback to remove bias.
However, such right censoring relies on the specific shape of the distribution,
and a missspecification will reintroduce bias.
Survival analysis also makes it harder to leverage available context when making
a decision: it either requires large amounts of data to fit a distribution for each
context, or fitting a more complex distribution that is more likely to
be misspecified.

\begin{figure}[t!]
    \begin{center}
      \includegraphics[width=0.75\columnwidth]{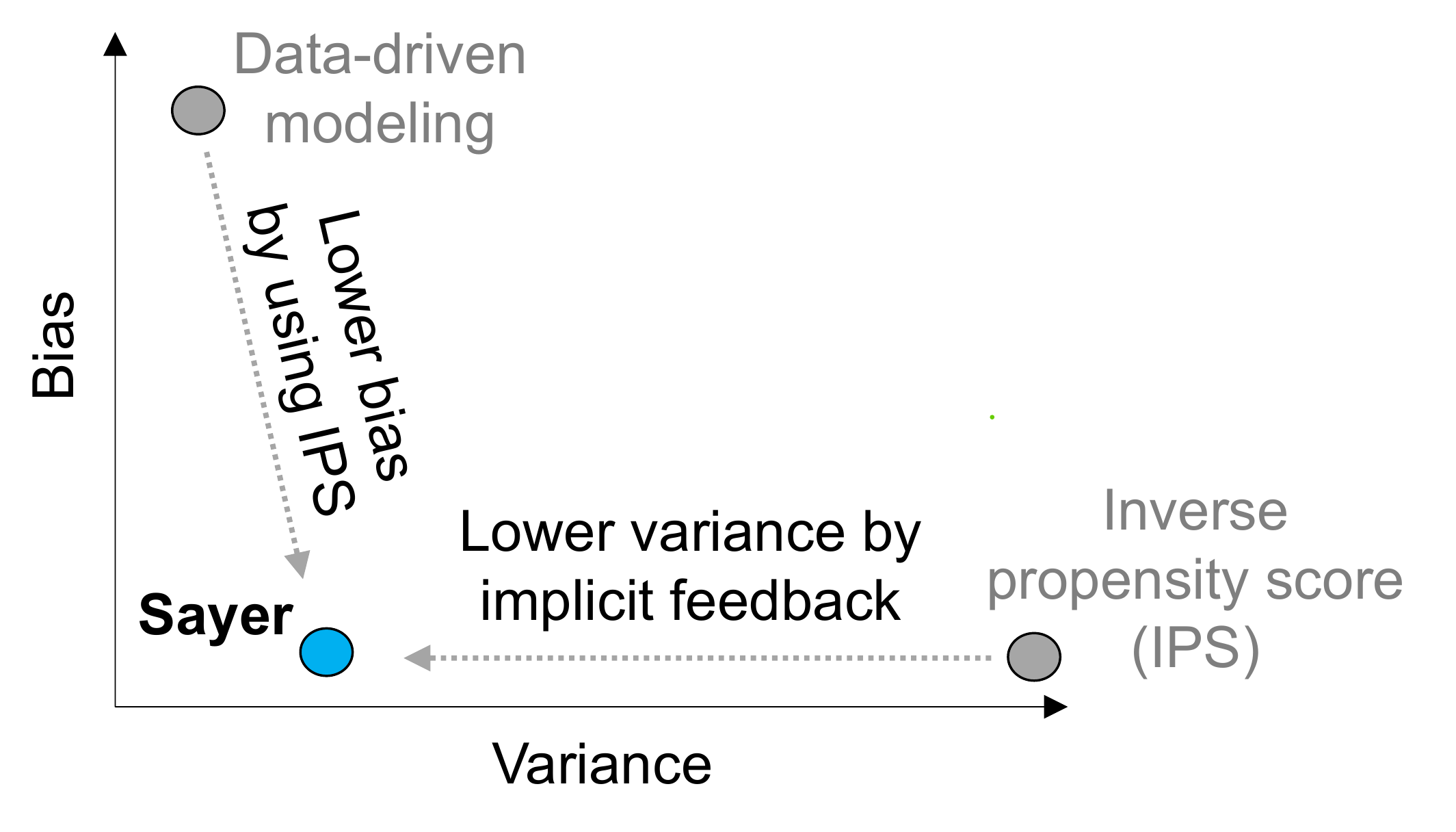}
      \tightcaption{{\bf \sysname vs. other approaches.} This figure is illustrative; actual figures appear later (\eg \ref{fig:counterfactual-barchart}).\vspace{-0.1cm}
      }
      \label{fig:contrast}
    \end{center}
\end{figure}

One way to avoid bias issues entirely is to use {\bf naive exploration}, in which the
deployed policy picks a random action for some fraction of the decisions, and the feedback
for these exploration decisions is used to evaluate candidate policies. For example,
\health currently waits the maximum time of 10 min for 35\% of its decisions (revealing
implicit feedback for all lower wait times), and the CFA system~\cite{jiang2016cfa} for
video QoE optimization collected data on a small portion of sessions by taking uniform random actions. 
These exploration datasets are unbiased and can hence be used to evaluate arbitrary
policies offline.

By additionally recording the probability of each chosen action, a technique called {\bf
Inverse Propensity Scoring (\ips)}~\cite{horvitz1952generalization} (and more advanced
variants like doubly robust estimators~\cite{dudik2014doubly}) can be used to leverage
both the exploration decisions and the biased decisions of the deployed policy.
This is because \ips uses probabilities to reweight (debias) the cost feedback
of each action to compensate for differences in the observed frequency of 
actions between the deployed policy and candidate policy. 
This reweighted data can be used to evaluate the cost of a candidate policy, or
compute updates to the policy during optimization.
Although the utility of \ips has been recognized~\cite{bartulovic2017biases,lecuyer2017harvesting}, 
it remains underutilized by the systems community. 
One reason for this may be that \ips requires a policy's decisions to satisfy
certain independence assumptions. Fortunately, these assumptions are naturally
satisfied by the threshold decisions we study.

We note that Table~\ref{tbl:approaches} identifies general properties that may not hold in
all situations. For example, if a threshold decision can be perfectly modeled, then a
simulator or data-driven model may be unbiased.
Since \ips comes closest to our goal, \sysname builds on \ips
to develop a methodology that is unbiased and data-efficient.
Figure~\ref{fig:contrast} illustrates the expected contrast between \sysname and these prior
methods.

\subsection{Goals of \sysname}
\label{subsec:goals}

As explained above, randomized exploration and \ips provide a good foundation for \sysname,
because they enable unbiased counterfactual evaluation of arbitrary policies. However,
these approaches suffer from a serious limitation: they only consider feedback for the (single) action 
taken by a policy decision. To leverage implicit feedback, \sysname must develop new techniques
that can be integrated easily into existing system policies. Specifically, \sysname addresses
two challenges:

\myparaitalic{{\em 1.} How do we leverage implicit feedback that is biased and outcome-dependent?} 
\sysname develops a new counterfactual estimator for implicit feedback (\S\ref{subsec:design:algorithm})
that is supported by new implicit exploration and data logging techniques (\S\ref{subsec:design:components}).

\myparaitalic{{\em 2.} How do we incorporate implicit feedback into the lifecycle of a
system policy?} \sysname modifies existing components of a system 
policy's workflow (\S\ref{subsec:design:overview}) while supporting its continuous optimization lifecycle 
(\S\ref{subsec:design:components}).

We focus on threshold decisions that satisfy the independence
assumptions mentioned above; for more complex decisions, \ips and \sysname may not be
appropriate (see \S\ref{sec:discussion}).
We evaluate \sysname in two production systems in \azure (\S\ref{sec:health},\S\ref{sec:scale}) using real production data and prototypes that mimic the production systems.


\section{Design of \sysname}
\label{sec:design}

This section presents the design of \sysname.
We state \sysname's assumptions and overview its architecture, 
and then present the key technical idea that
enables unbiased counterfactual evaluation based on implicit feedback. We then show how \sysname
integrates with the lifecycle of a system policy.

\subsection{Overview}
\label{subsec:design:overview}

\mypara{Terminology and assumptions} 
A {\em system policy} ($\pi$) makes decisions by taking the {\em context} ($\vec{x}$) of a decision as
input and choosing a single {\em action} ($\pi(\vec{x})$) to take from a set of allowed
actions.
The context $\vec{x}$ comprises properties of the environment or the system state that are
considered relevant to the decision.
Traditionally, when the policy takes an action, we observe the {\em cost}
({\em feedback}) $c(\pi(\vec{x}))$ associated with that action.
But as observed in \S\ref{subsec:implicit}, in system policies that make threshold
decisions, we can deduce the cost of more actions than the one we actually take, \ie implicit feedback.

We make three assumptions that are relevant to the system policies we study.
First, we assume that the decisions made by a policy are mutually independent.
This assumption corresponds to the {\em contextual bandits}
setting~\cite{langford-nips07,learning-from-ips} and is required by the \ips
estimator we build on.
Intuitively, it means that the action chosen by the system policy at a given
time does not influence the cost of future actions.
For instance, such a long-term influence could arise if an action drastically
changes the load of the system, thereby changing the future distribution of
contexts; or by changing the state of a cache, thereby changing the cost of a
future action with the same context.
Fortunately, the independence assumption is a good model for the threshold
decisions in \S\ref{subsec:implicit}, as they involve one-step decisions in
large enough systems that the impact of individual decisions are well isolated.

Second, we assume the policy makes a one-dimensional, discrete decision. That
is, the actions are the (multiple) possible values for a single parameter.
This is a standard assumption in the learning literature we build on, where
exploring continuous, unbounded, or complex action spaces quickly becomes
intractable without strong structure~\cite{slates}.
When the action space is continuous, as in our \health example, we
can discretize the possible actions within a chosen range, \eg between 0 and 
a maximum wait time fixed a priori.
This restriction does not apply to the context or cost function, which
may be continuous and multi-dimensional.

Third, we assume that the observed potential outcomes do not depend on the chosen action.
This assumptions is a direct extension of the inclusion restriction assumption from causal inference \cite{imbens-rubin-causality} to our implicit feedback setting.
More concretely, it means that in \health, the recovery state of a machine 
after waiting for $t$ does not depend on whether the reboot timeout is set to $t+1$ or $t+10$.
This is a natural assumption, because the timeout is never acted on until the actual reboot.
In the \scale application, though, over-allocating by ten or twenty VMs could yield different completion times 
for a particular VM (\eg due to queueing effects), violating the assumption.
Fortunately, we verify that the assumption holds in practice (\S\ref{sec:scale}, Figure \ref{fig:causal-assumption-vmss}).

Finally, the contextual bandit literature typically also assumes that contexts and costs come from stationary distributions.
This enables bandit algorithms to gradually learn the distribution and decrease exploration over time.
However, practical system deployments often experience context and cost distributions that change over time.
In this work, we thus focus on continuous exploration, which allows us to cope with this kind of non-stationarity.

\begin{figure}[t]
    \begin{center}
      \includegraphics[width=0.8\columnwidth]{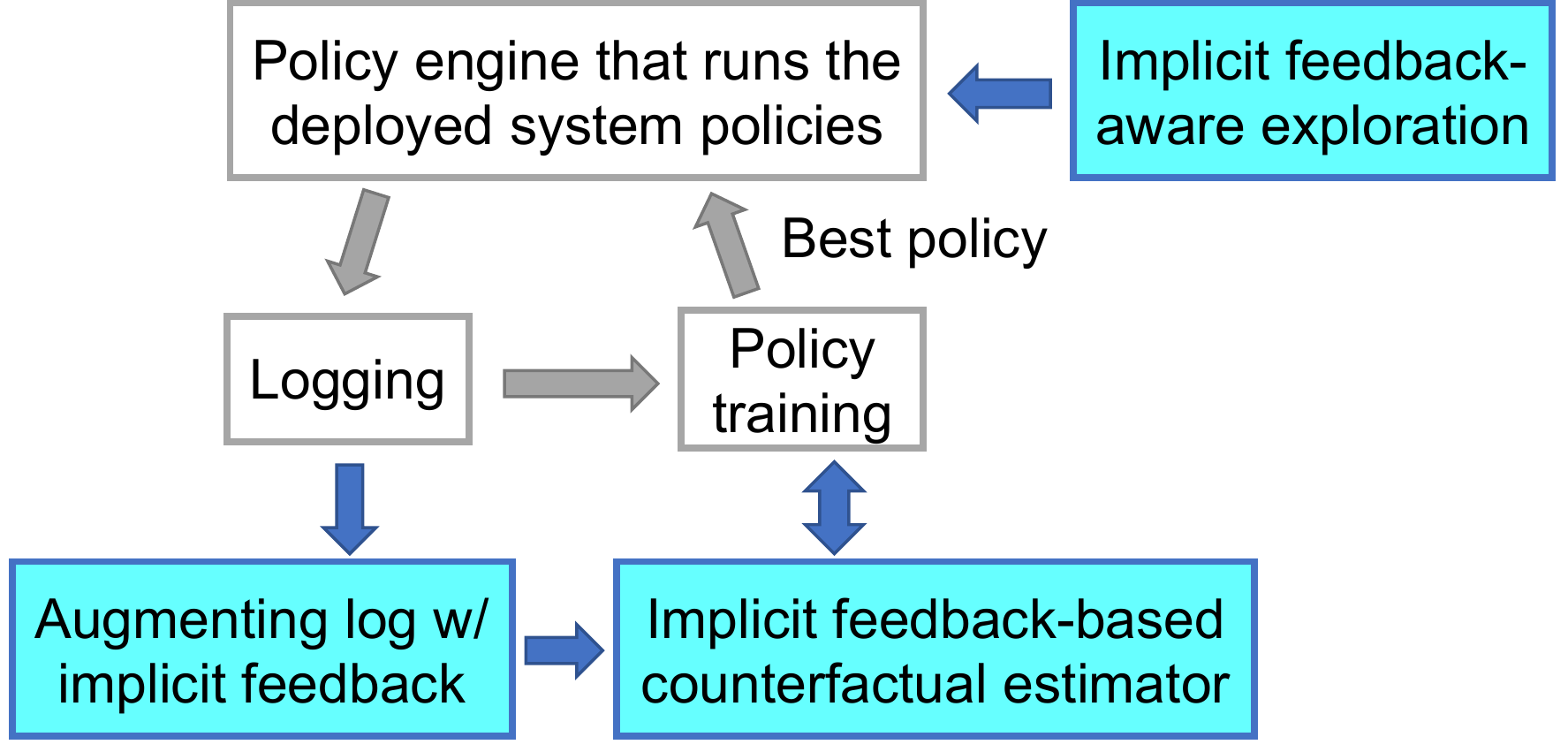}
      \tightcaption{
{\bf Architecture of \sysname.} \sysname expands the traditional policy optimization workflow by adding three new modules
(highlighted in blue) that leverage the available implicit feedback. 
}
      \label{fig:arch-new}
    \end{center}
\end{figure}

\mypara{Architecture}
Figure~\ref{fig:arch-new} shows the architecture of \sysname. 
The gray boxes are standard components of a decision-making system, which include: a policy engine that hosts the
deployed policy and invokes it on every decision to obtain the policy's chosen action, a logging component that 
records the outcome (cost) of each decision, and a policy training component that updates the policy based on the
logged data. 
\sysname adds the outlined blue boxes, which we describe in detail in \S\ref{subsec:design:components}.
The added components can be integrated seamlessly into an existing system. 
For example, \sysname's implicit exploration adds some randomization to the deployed policy's decisions,
but this does not change the policy deployment interface. 
Similarly, existing logging components typically already log extensive information about each 
decision (\eg context, action, and cost); \sysname additionally logs information about the
randomization added by the implicit exploration component (in the form of action probabilities).
\sysname uses this information to augment the traces output by the logging component with implicit feedback,
which is done as a separate post-processing step.
 
The key component of \sysname is an unbiased counterfactual estimator that uses the implicit feedback augmented 
in the logged data of the deployed policy, to estimate the cost of an arbitrary candidate policy 
on the same sequence of decisions. 

\subsection{Implicit feedback-based counterfactual estimator}
\label{subsec:design:algorithm}

We present our algorithm for implicit feedback-based counterfactual estimation in a progressive
fashion, starting from a basic framework that guarantees unbiased estimation.

In the simplest setting where a policy only receives feedback for the single chosen
action, also called {\em bandit feedback},
the data collected will be biased. Here, \ips provides a framework for unbiased estimation.
 
\mypara{\ips: Unbiased estimator for bandit feedback}
To remove bias from a log of bandit feedback data, a classic solution is to log the probability ($p$) of the action 
being chosen alongside the context, action, and cost of the decision, yielding a tuple \tuplerl. Based on this log, a technique called {\em Inverse Propensity Scoring}
(\ips)~\cite{horvitz1952generalization} can then provide an unbiased estimator for any candidate policy
$\pi$:
\begin{align}
  \ips(\pi) = \textstyle \tfrac{1}{N}~\sum_{\substack{\tuplerl}}
  \frac{\1\{ \pi(\vec{x}) = a \}}{p} c . \nonumber
\end{align}
\ips finds instances in the trace where $\pi$ chooses the same action as the deployed policy, \ie $\pi(\vec{x}) = a$ (the indicator function $\1$ has value $1$ under a match and $0$ otherwise).
But instead of simply adding the associated cost under a match, it re-weights it by the inverse of the probability ($p$) that $a$ was chosen.  
Thus if $a$ has low probability, \ie it is rarely chosen by the deployed policy, \ips will upweight it to compensate for the fact that $a$ is underrepresented in the trace.

\begin{figure}[t]
    \begin{center}
      \includegraphics[width=0.99\columnwidth]{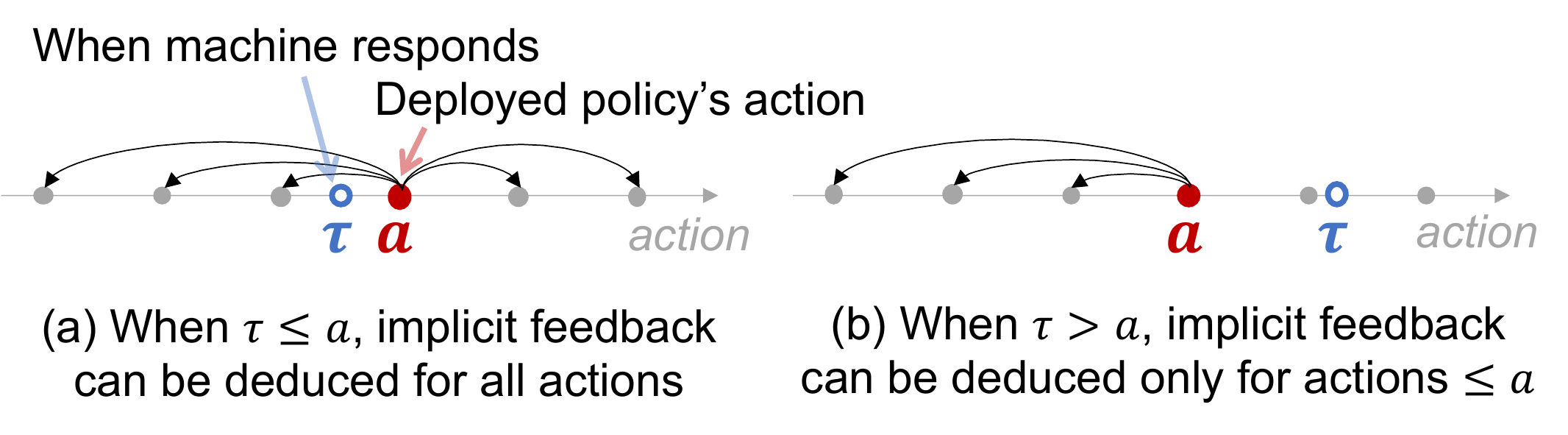}
      \tightcaption{{\bf Implicit feedback example.} 
      Each black arrow indicates that the feedback of an action can be deduced from the
      feedback of the deployed policy's action. Importantly, implicit feedback is {\em
      outcome-dependent}: if a machine responds (at $\tau$) before
      the chosen wait time $a$, we obtain full feedback, but otherwise we only obtain
      feedback for actions $\le a$.
}
      \label{fig:feedbackgraph}
    \end{center}
\end{figure}

\mypara{Implicit feedback vs. bandit feedback}
\ips is an unbiased estimator for bandit feedback, but it completely ignores any feedback that may be received for actions other than $a$, \ie implicit feedback. 
Implicit feedback thus provides a type of partial feedback that lies between full feedback and bandit feedback.
Although partial feedback has previously been studied using feedback graphs~\cite{mannor-feedback,alon-feedback}, those algorithms take a fixed feedback graph as input and optimize one policy online.
In contrast, we are concerned with counterfactual evaluation of many policies, and the feedback we receive {\em depends on the outcome of each decision}.
For example, in \health, if a machine responds
at time $\tau$ before the chosen wait time $a$, we obtain full feedback, but if it does not,
we only obtain feedback for actions $\le a$.
This is illustrated in Figure~\ref{fig:feedbackgraph}.
Since \ips requires a probability $p$ that is fixed in advance, it does not support such variable feedback out of the box.
%

So how do we incorporate implicit feedback into \ips without compromising its unbiasedness?

\mypara{\implicit: Unbiased estimator for implicit feedback} 
Our key insight is that \ips can be interpreted as matching datapoints in the trace
according to an {\em event} $E$ under which we (1) know the cost feedback and (2) can compute the
probability that $E$ occurs, $P(E)$, in order to reweight the cost.
We can thus abstract \ips to obtain a template for our estimator:
\begin{align}
  \implicit(\pi) = \textstyle \tfrac{1}{N}~\sum_{\substack{\tupleim}}
  \frac{\1\{E\}}{P(E)} c(\pi(\vec{x})) . \nonumber
\end{align}

In the original \ips estimator, $E=\{\pi(\vec{x}) =
a\}$ is the event that the chosen actions match. $P(E) = p$ since the actions match
precisely when the deployed policy chooses $a$.

Now, by redefining $E$ to be a larger event, we can match a larger set of points,
while preserving the unbiasedness of this template! In particular, \sysname defines $E$
as the event that ``the feedback of the candidate policy's action $\pi(\vec{x})$ 
can be deduced from the outcome of the deployed policy's action $a$''.
Intuitively, $\1\{E\}$ is 1 only if the feedback $c(\pi(\vec{x}))$ is available, either
explicitly through matching, or implicitly through deduction from the outcome of action
$a$ recorded in the trace. Figure~\ref{fig:feedbackgraph} illustrates an example.
$P(E)$ is then the probability that the deployed policy chooses an action whose outcome will
allow the feedback $c(\pi(\vec{x}))$ to be deduced.

To make this more concrete, we derive $E$ and $P(E)$ for \health. 
The application to \scale is similar.

\mypara{Applying \implicit to \health}
We define $E$ based on a given chosen action (wait time) $a$, its outcome (either the
machine responds at time $\tau \le a$, or it times out), and a new action chosen by the
candidate policy, $\pi(\vec{x})$.
The event that $\pi(\vec{x})$'s feedback can be deduced is: \[
  E = \{\pi(\vec{x}) \le a\} \cup \{ \tau \le a\} .
\] The first term follows from the property of \health's decisions that when $\pi(\vec{x})
\le a$, $c(\pi(\vec{x}))$ can always be deduced (see Figure~\ref{fig:feedbackgraph}).
Otherwise, $c(\pi(\vec{x}))$ can be deduced only when $\tau \le a$ 
(Figure~\ref{fig:feedbackgraph}(a)).
In other words, the only case when feedback is not available is when the deployed
policy's action $a$ causes a timeout (\ie $\tau>a$) and the candidate policy chooses to
wait even longer $\pi(\vec{x}) > a$. The recorded outcome does not provide any
information in this case.

\begin{figure}[t]
    \begin{center}
      \includegraphics[width=0.99\columnwidth]{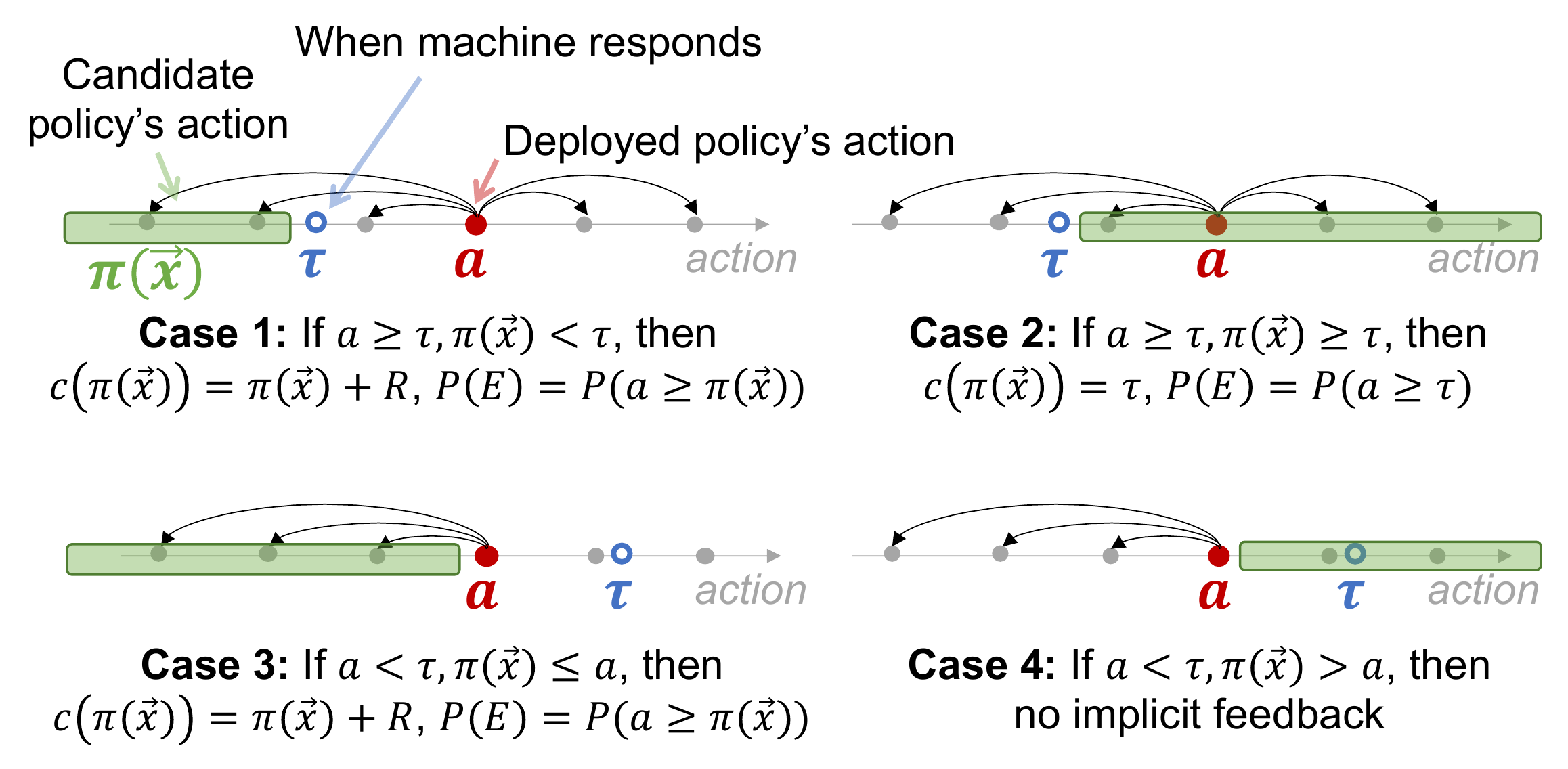}
      \tightcaption{Example of how to calculate $c(\pi(\vec{x}))$ and $P(E)$ based on implicit feedback in \health. }
      \label{fig:cases}
    \end{center}
\end{figure}

Next, whenever $\1\{E\} = 1$, we need to compute $P(E)$ as well as the cost of the candidate policy's action $c(\pi(\vec{x}))$. 
There are four cases, as illustrated in Figure~\ref{fig:cases}.
\begin{packeditemize}
\item {\bf Case 1:} $a \geq \tau$ and $\tau > \pi(\vec{x})$. In this case, $\pi(\vec{x})$ will cause the machine to reboot, so $c(\pi(\vec{x})) = \pi(\vec{x}) + R$, where $R$ is the the fixed reboot delay cost (see \S\ref{sec:health} for details).
  Moreover, this information is available if and only if the deployed policy chooses an action $\geq \pi(\vec{x})$, so $P(E) = P(a\geq\pi(\vec{x}))=\sum_{a' \geq \pi(\vec{x})} p(a')$.
\item {\bf Case 2:} $a \geq \tau$ and  $\pi(\vec{x}) \geq \tau$, which means both deployed and candidate policies will wait long enough for the machine to respond, 
so $c(\pi(\vec{x})) = \tau$.
This information is available if and only if the deployed policy chooses an action $\geq \tau$, so $P(E) = P(a\geq\tau)=\sum_{a' \geq \tau} p(a')$.
\item {\bf Case 3:} $\tau > a$ and $a \geq \pi(\vec{x})$,  which means that the deployed policy's action is not long enough for the machine to respond before rebooting, and the candidate policy chooses to wait even less time. 
So $c(\pi(\vec{x})) = \pi(\vec{x}) + R$.
This information is available if and only if the deployed policy's action is greater than the candidate policy's action, so $P(E) = P(a\geq\pi(\vec{x}))=\sum_{a' \geq \pi(\vec{x})} p(a')$.
\item {\bf Case 4:} $\pi(\vec{x}) > a$ and $\tau > a$, which means $E$ is false and $\1\{E\}=0$, so we do not need to compute anything. 
\end{packeditemize}

\mypara{Unbiasedness and low variance of \ \implicit}
\implicit's unbiasedness follows directly from that of \ips.
Given a datapoint $(\vec{x}, \vec{c})$ and the action $a$ chosen by the policy that was deployed at the time, we have:
\begin{align}
  \Exp\Big( \frac{\1\{E\}}{P(E)} c(\pi(\vec{x})) \Big) = \frac{\cancel{\Exp\big(\1\{E\}\big)}}{\cancel{P(E)}} c(\pi(\vec{x})), \nonumber
\end{align}
implying that $\Exp\big(\implicit(\pi)\big) = c(\pi(\vec{x}))$, and hence that \implicit is an unbiased estimate of the cost of policy $\pi$, had it run on the same sequence of datapoints observed during data collection.
We can see that $P(E)$, which we call the {\em implicit probability}, plays a key role in this unbiasedness guarantee.
Indeed, reweighting the implicit feedback deduced from matched events by $1/P(E)$, similar to \ips's reweighting by $1/p$, cancels out the $\Exp\big(\1\{E\}\big)$ term in the expectation, thereby removing the bias due to missing information.

In addition, $P(E)$ lowers the variance of \implicit's estimates, as compared to \ips.
Computing the variance yields:
\begin{equation}
  \label{eq:implicit-var}
  \Var\Big( \frac{\1\{E\}}{P(E)} c(\pi(\vec{x})) \Big) = \frac{1-P(E)}{P(E)} c(\pi(\vec{x}))^2 .
\end{equation}
As we can see, the variance of the cost is scaled by $1/P(E)$.
In \ips, the weight $1/p$ can be quite high when the deployed policy and candidate policy differ.
In contrast, \implicit matches a range of actions and $P(E)$ is their total probability; this leads to larger values of $P(E)$ and hence lower variance.

\subsection{Implicit feedback-based policy optimization}
\label{subsec:design:components}

We now discuss the system components of \sysname (Figure~\ref{fig:arch-new}),
which integrate the implicit counterfactual estimator above
into the lifecycle of system policy optimization.

\mypara{Implicit data augmentation}
\sysname's logging component ensures that the information of each decision is recorded correctly, so that the generated traces can be used for counterfactual evaluation and training. 
In a standard bandit feedback setting, the logging component would log the tuple \tuplerl, where $\vec{x}$ is the input context, $a$ the chosen action, $c$ the cost feedback received for $a$, and $p$ the probability of choosing $a$.  
In \sysname, since there is implicit feedback, we additionally log the implicit probability and cost of all actions 
whose feedback can be deduced, \ie $(\vec{x},\{(a_i,c_i,p_i)\}_i)$.
This information is used by our implicit counterfactual estimator to correctly handle implicit feedback.
In \health, for example, if a logged action $a=5$~min leads to a machine responding
at $\tau = 3$~min, then instead of logging $(\vec{x}, 5~\textrm{min},
3~\textrm{min}, p)$, we augment the entry with implicit feedback (using Case 1 and Case 2 in
Figure~\ref{fig:cases}): $(\vec{x}, \{(a<3~\textrm{min}, (a+R)~\textrm{min},
P(a<3~\textrm{min})),(a\geq3~\textrm{min}, 3~\textrm{min}, P(a\geq3~\textrm{min}))\})$.

\mypara{Counterfactual estimation and policy training}
Given the logged data augmented with implicit feedback and probabilities, 
\sysname uses its implicit counterfactual estimator to evaluate and train new policies. 
To train a policy, \sysname uses a class of contextual bandit learning
algorithms that internally use estimators such as \ips to efficiently search a policy
space~\cite{learning-from-ips}, but replaces the internal estimator with \implicit.
\sysname follows the common practice of splitting the logged data into a train
set (to learn the new policy) and a test set (to evaluate it); the difference is
that \sysname uses {\em counterfactual} training and evaluation algorithms, to ensure
unbiasedness.
The granularity at which \sysname is run, and the subset of data used to train and test, 
is largely orthogonal to our work.
For example, in our evaluation we run \sysname in a ``batch retraining'' mode, where
a new policy is periodically trained from scratch on a trailing window of data,
before being counterfactually evaluated on the most recent data.

By repeating the above process, \sysname enables a continuous optimization loop
that allows a system policy to evolve. Continuous optimization is necessary to
cope with changes or non-stationarity in the system or environment
(\S\ref{sec:health},\S\ref{sec:scale}).

\mypara{Implicit exploration and logging}
To enable unbiased counterfactual estimation, and to cope with non-stationarity, \sysname
includes an exploration component that adds controlled randomization to the
deployed policy's decisions. This ensures that feedback is received for all actions, not
just those deemed optimal by the deployed policy (the classic exploration-exploitation tradeoff).
A simple algorithm for exploration is \epsgreedy~\cite{langford-nips07},
which selects a random action \eps fraction of the time and uses the action chosen by the
deployed policy the remaining $1-\eps$ fraction of the time.
 
Randomizing over all actions makes sense when feedback is only received for the chosen
action (\ie bandit feedback).
However, this means that the probability of choosing an action can be as low as
$p = \epsilon/a$, yielding high variance (see Equation \ref{eq:implicit-var}) even with
a large fraction of exploration actions.
In our setting, where we receive implicit feedback for other actions, 
randomizing over all actions ``underutilizes'' this feedback.
Instead, \sysname \ {\em explores by selecting the maximal action}, \ie the action that
yields the most feedback.
For example, in \health, this corresponds to waiting 10 min in each explore step.
Using the maximal action for exploration means that the minimal observation
probability is $P(E) = \epsilon$, reducing the variance of our counterfactual
estimator and supporting smaller amounts of exploration. Thus for a given
exploration budget, \sysname's implicit exploration technique better utilizes
implicit feedback.
This is particularly important to enable continuous exploration at low cost.

Of course, exploring using the maximal action may not always be desirable, especially
if that action is likely to be costly.
We use this heuristic because it allows us to explore less frequently, and because in
our applications the maximal action is a reasonable default
action that we are trying to improve upon.
An alternative would be to explore randomly over the action space using a
distribution that accounts for the desirability of each action. However,
this would require more frequent exploration, and would not completely avoid 
exploring the maximal action, which is necessary to collect
unbiased data.

\newcommand{\direct}{\textsf{Direct Method}\xspace}
\newcommand{\survival}{\textsf{Survival Analysis}\xspace}
\newcommand{\naive}{\textsf{Naive Implicit}\xspace}
\newcommand{\full}{\textsf{v0}\xspace}
\newcommand{\oneshot}{\textsf{v1}\xspace}
\newcommand{\omniscient}{\textsf{Omniscient}\xspace}

\section{Evaluation}

We now demonstrate the value of \sysname in two real applications from \azure---\health and \scale. 
Using a combination of simulation (driven by full-feedback traces and synthetically generated traces) and A/B testing in a live prototype, we show that 
{\em (i)} compared to baseline performance estimators, \sysname's \implicit counterfactual estimator is unbiased and has lower variance;
and {\em (ii)} the more accurate counterfactual estimation allows \sysname to train better
system policies than the baselines, even though \sysname incurs the cost of exploration.

\subsection{Methodology}
\label{sec:eval-methodology}

We define the baselines and performance metrics common to both applications here. 
In \S\ref{sec:health} and \S\ref{sec:scale}, we introduce the details of each application.

\mypara{Baseline estimators} 
Given a log of \tuplerl tuples, we consider four baseline performance estimators for counterfactual evaluation and optimization:
\begin{itemize}
\item {\em \direct}  uses the	log to train a linear model using Vowpal Wabbit (\vw)~\cite{vw}, a state-of-the-art bandit library, that predicts the cost of any context and action.
This can be viewed as a data-driven performance modeling baseline, which may be biased by the log and the model.
The cost model is used to predict missing information when doing counterfactual evaluation, and when training a new policy.
\item {\em \survival} postulates a distribution over the outcomes for all decisions (\eg the distribution of machine recovery times).
Like the \health team, we fit a right-censored Lomax (long tail) distribution using maximum likelihood estimation (implemented with Pyro \cite{bingham2018pyro}).
Right censoring accounts for unobserved costs by putting all the probability mass of the tail on the maximum observed value.
We fit this distribution on all observations, since conditioning on features reduces the amount of data and yields poor performance.
Missing costs are replaced with their expected value for counterfactual evaluation, and the learned policy is the action with the lowest expected cost.
\item {\em \ips} is described in \S\ref{subsec:design:algorithm}.
This classic counterfactual estimator is well known to be unbiased, but has high variance
and does not leverage implicit feedback.
It is used both for counterfactual estimation, and to debias the cost during policy optimization using \vw~\cite{vw}.
\item {\em \naive} is the \direct trained on the log augmented with implicit feedback (like \sysname), but without using implicit probabilities.
It can be viewed as using implicit feedback without proper reweighting, and is used to emphasize the importance of reweighting to remove bias.
%
\end{itemize}

\mypara{Baseline policies} 
To ensure fairness, all policies use the {\em same} training interface of \vw (which accepts as input a log of \tuplerl tuples augmented by each estimator) and train the {\em same} model (a linear contextual bandit policy)---except \survival, which is a different type of model.
This process yields five policies: \{~\sysname, \direct, \survival, \ips, \naive\}-based policies.
These policies use different exploration strategies: \direct always uses the action
returned by the trained policy; for a random $\epsilon$ fraction of decisions, \sysname
and \naive use the maximal, full-feedback action (since both use implicit feedback), while \ips uses
a random action (since it only matches the exact action).

We also include two baselines that are not continuously updated, but serve as useful reference design points:
\begin{itemize}
\item {\em Full-feedback (\full)} policy chooses the most conservative action and obtains full feedback. 
It maximizes information in the collected data at the cost of performance.
\item {\em One-shot learning (\oneshot)} policy feeds the full-feedback data (from which we can deduce the true outcome of every action) to \vw to train a policy, but never updates the policy. 
\end{itemize}

Finally, a skyline policy shows an upper bound on the performance one can hope to achieve with our \vw model:
\begin{itemize}
\item {\em  \omniscient} is a policy trained on full-feedback data, even when such
feedback would not be available from the data collection. This is only available in setups
where we have access to full feedback, and simulate partial feedback by hiding information.
\end{itemize}

\mypara{Metrics} 
We compare these approaches under a continuous optimization scenario, in which trained policies are deployed, 
and updated using a rolling window of data collected while they were running.
We focus on the following metrics:
\begin{itemize}
\item {\em Trained policy cost:} Each system defines a cost associated with policy actions. 
We measure the cost of running a trained policy (lower is better).
\item {\em Counterfactual evaluation error:} Using data collected when running the policy, we measure the difference between the estimated cost for a different candidate policy, and the true value for this cost.
\end{itemize}
For each metric we show both the {\em average performance} (cost), as well as the {\em range of values} that can happen due to randomness in the data.  
These ranges  are computed using bootstrap~\cite{efron1979bootstrap}, a common resampling
technique from statistics analogous to averaging over repeated experiments, which gives a sense of the variance in our results.

\subsection{Application I: \health}
\label{sec:health}

\health is a monitoring service that uses heartbeats to detect unresponsive physical machines within datacenters, and is responsible for rebooting unresponsive machines after a threshold amount of time. 
We formalize the problem as follows: faced with an unresponsive physical machine, the policy chooses a wait time amongst ten options (the action set) \{1,2,...,10\} min, before rebooting the machine.
This maximum wait of 10 minutes is a practical limit imposed by the \health team, regardless of any RL constraints.
The decision is based on a number of context features for the machine, collected via an existing telemetry pipeline, and available to the policy at decision time.\footnote{In this setting, the \direct resembles a heuristic predicting the downtime of stragglers based on the history of similar machines, \eg~\cite{ananthanarayanan2013effective,zaharia2008improving}.}
This context includes the hardware/OS configuration of the machine, which cluster it belongs to, the number of previous failures in the cluster, and the number of client VMs running on the machine.
The cost of an action $a$ is the total downtime experienced by customer VMs.
It is calculated as the recovery time $\tau$, plus a fixed reboot time $R=10~\textrm{min}$ if the server is rebooted  ($a<\tau$),
scaled by the number of VMs on the server $N_{\textrm{VMs}}$:
\[
  cost = N_{\textrm{VMs}} \Big(\1\{a < \tau\} \tau + (1 - \1\{a < \tau\}) (a + R)\Big)
\]
As explained in \S\ref{subsec:design:algorithm}, \health provides implicit feedback. 
Choosing action $a$ reveals the cost of all actions $a' \leq a$, as we know the corresponding state of the machine. If we observe a recovery ($a \geq \tau$), we can also deduce the cost of every action, making it full feedback.

\begin{figure}[!t]
  \centering
    \includegraphics[width=0.9\linewidth]{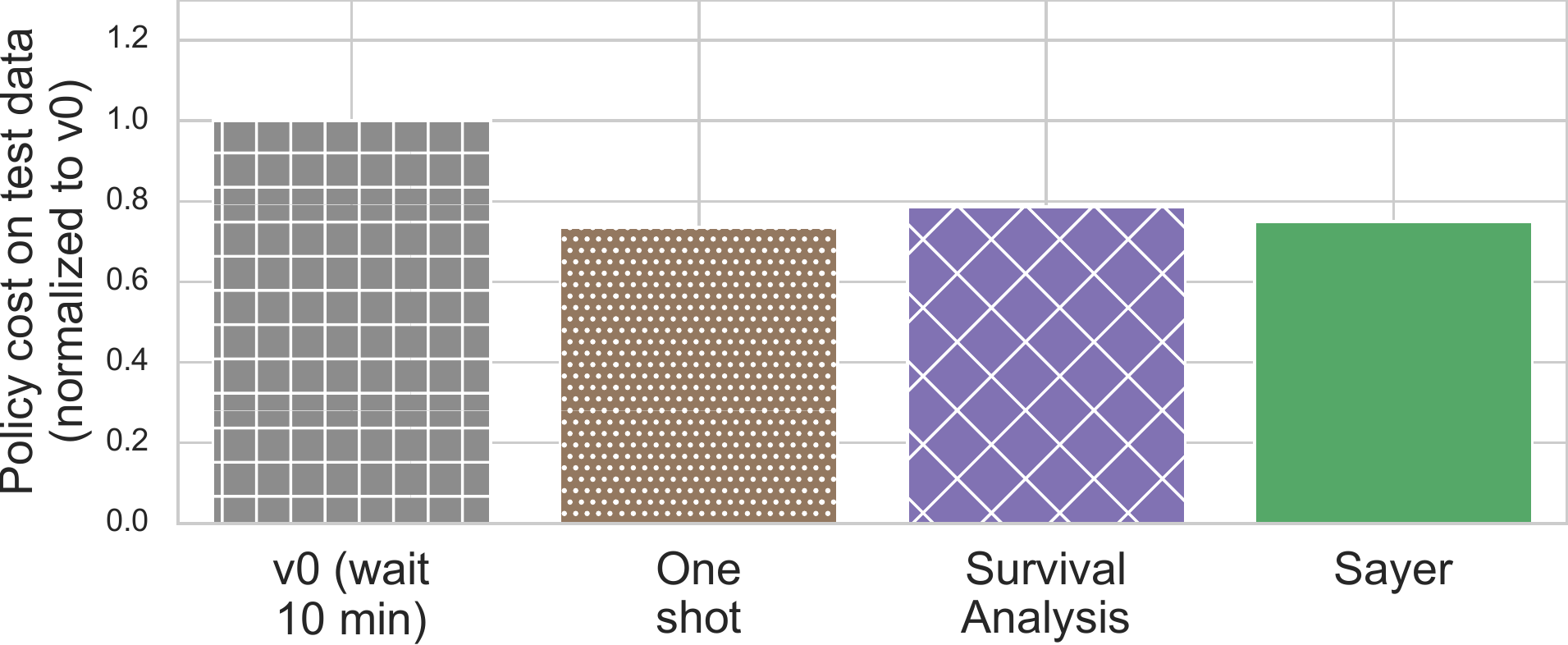}
  \tightcaption{{\bf First training period:}
  We use the 2nd split (\stwo) to
  compare the performance of policies trained on the 1st split (\sone), relative to the
  \full policy. 
  }
  \label{fig:s1}
\end{figure}

\begin{figure*}[!ht]
  \centering
  \begin{subfigure}{.49\textwidth}
    \centering
\includegraphics[width=0.99\linewidth]{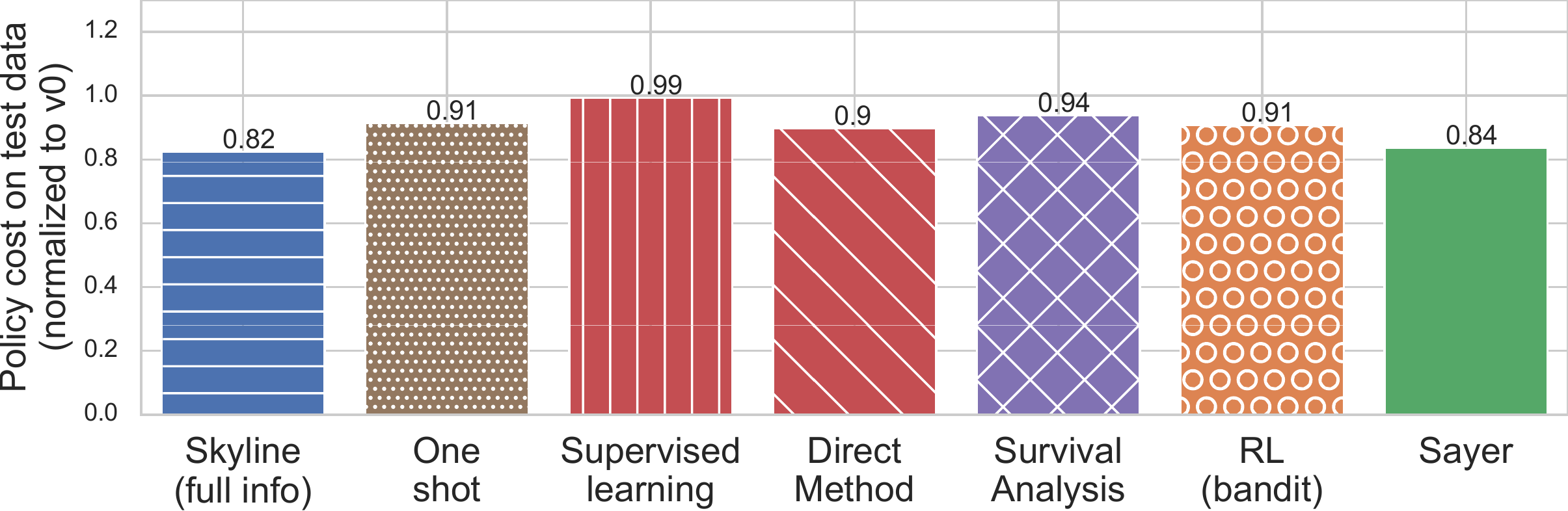}
    \caption{\footnotesize {\bf Cost of policies trained by various estimators (x-axis)}. 
    }
    \label{fig:optim-barchart}
  \end{subfigure}%
  \hfill
  \begin{subfigure}{.49\textwidth}
    \centering
\includegraphics[width=0.99\linewidth]{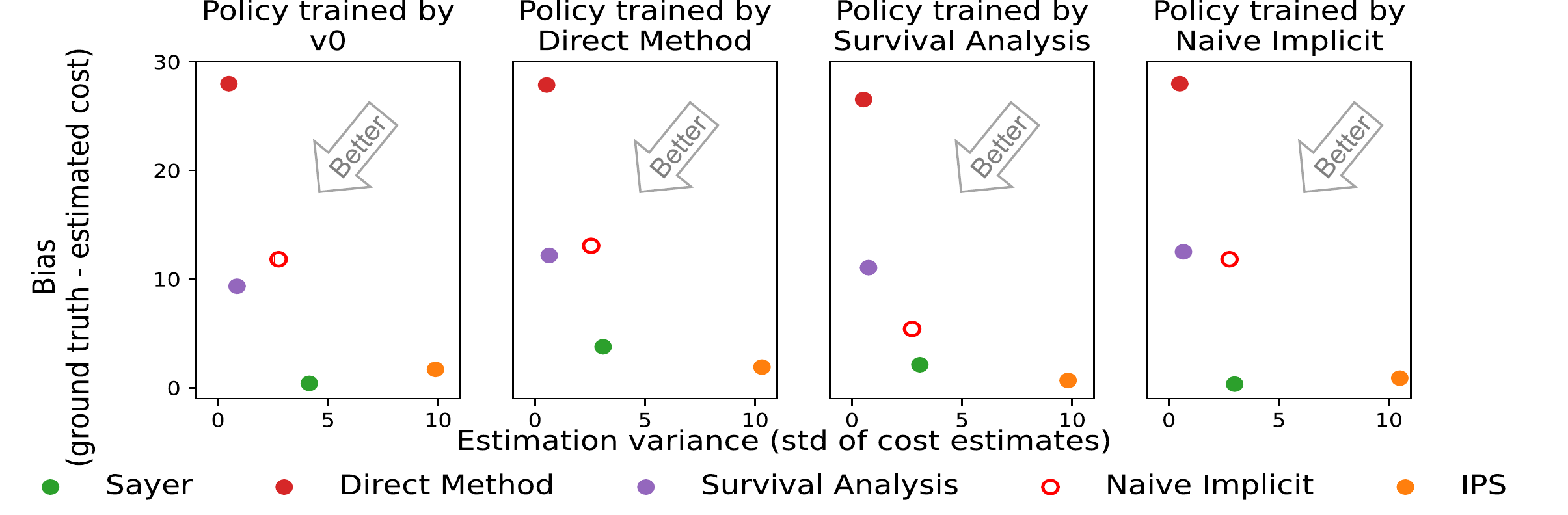}
\caption{\footnotesize {\bf Counterfactual estimation error (bias \& variance) on \ref{fig:optim-barchart} policies}. 
}
    \label{fig:counterfactual-barchart}
  \end{subfigure}
  \caption{Policy optimization and evaluation over time.
  All plots show costs (downtime) normalized to
  the cost of the default policy (\full). 
  (\ref{fig:optim-barchart}) shows the performance of policies retrained
  from data they generated in \stwo, when evaluated on \sthree.
  (\ref{fig:counterfactual-barchart}) shows the counterfactual
  estimate (one colored dot per estimator) of how the policies from \ref{fig:optim-barchart} (one per box) would
  perform on \sthree, using data collected by deploying the \oneshot policy,
  normalized to the full-feedback ground truth.
  \sysname achieves both unbiased and low variance cost estimates, which trains better policies (\ref{fig:optim-barchart}).
  }
  \label{fig:health-barcharts}
\end{figure*}

\subsubsection{Trace-driven simulation}
\label{s:eval-health-trace}
~

We obtained a production trace of $13.5k$ events with full feedback from \health, collected
during an initial two-month period when the team deployed the conservative policy of always waiting the maximum of 10 min.
Based on this trace, we can compute the ground truth performance of different policies, and thereby evaluate the performance of \sysname's counterfactual estimator and its trained policies.
We split the production trace into three periods, each with 3000-5000 data points, corresponding to 
environmental changes:
\begin{itemize}
\item \sone corresponds to the phasing out of a hardware configuration (a specific server SKU), 
\item \stwo lies between the events of \sone and \sthree, and 
\item \sthree corresponds to a major software upgrade. 
\end{itemize}
Because these hardware and software configuration changes affect large portions of the machines, we also expect them to affect the machine's recovery times and the relationship between observed features and recovery times, which we observe in the data.
We leverage these three phases to evaluate how different approaches cope with environmental changes compared to \sysname.

\mypara{First training period}
First, without any prior knowledge, we start with \full (choosing the maximal wait time of 10 minutes) in \sone, which produces full feedback, train new policies according to the different approaches described in \S\ref{sec:eval-methodology}, and evaluate these policies in \stwo.
Figure \ref{fig:s1} shows the results. 
The policies trained by estimators \direct, \naive, and \oneshot will be identical on full feedback. 
They yield a $26.5\%$ improvement over the \full baseline policy.
On \sone, \sysname will learn the same policy, but is slightly less efficient because of the added exploration, which chooses the maximal action on $10\%$ of the actions. 
Despite this exploration, \sysname yields a $25\%$ improvement over the \full baseline.
\survival is less expressive, as it chooses a single wait time for all machines, but it still yields a $21\%$ improvement 
(alternatively, we could train one model per cluster to take features into account, but this degrades the results due to the small data).

\mypara{Continuous training}
Second, we evaluate different policy training approaches in a dynamic setting by deploying each policy on \stwo (yielding the performance from Figure~\ref{fig:s1}), collecting the resulting feedback, and retraining each policy on the data generated while it was running.
For instance, if a policy waits $3$ min for a machine, we only show the costs for actions $a \leq 3$ in case of a reboot, and all costs if the machine recovers within that time frame.
This partial feedback data is used to train the new policy, which is then
evaluated on \sthree.
Figure~\ref{fig:optim-barchart} shows the policy cost for each of these
retrained policies, measured on \sthree using full feedback.
There are several observations about \sysname's performance:
\begin{itemize}
\item 
{\em Substantial benefits by continuous retraining:}
As we see in Figure \ref{fig:optim-barchart},
the \omniscient skyline policy, which trains its policy on \stwo with full feedback,
yields a $15.5\%$ improvement over \full when deployed on \sthree.
However, keeping the one shot model trained on \sone only gives an $8.6\%$ improvement.
We expect even larger degradation as the environment evolves.

\item 
{\em Benefits of implicit exploration:}
Naively retraining on data collected when running an optimized policy (\ie without exploration) is suboptimal. 
Figure~\ref{fig:optim-barchart} shows that training a policy with \naive, which ignores missing feedback from a lack of exploration, is almost as bad as \full.
Even extrapolating missing feedback using \direct or \survival (which does improve performance to $7.6\%$ and $6.1\%$, respectively) is still short of the \oneshot baseline (trained on older but full feedback data).
\sysname is the only one to reach \omniscient's performance, 
again with a small added cost for exploration, yielding a $14.7\%$ improvement.
\end{itemize}

\begin{figure*}[t]
  \centering
  \begin{subfigure}{.57\textwidth}
    \centering
    \includegraphics[width=0.7\linewidth]{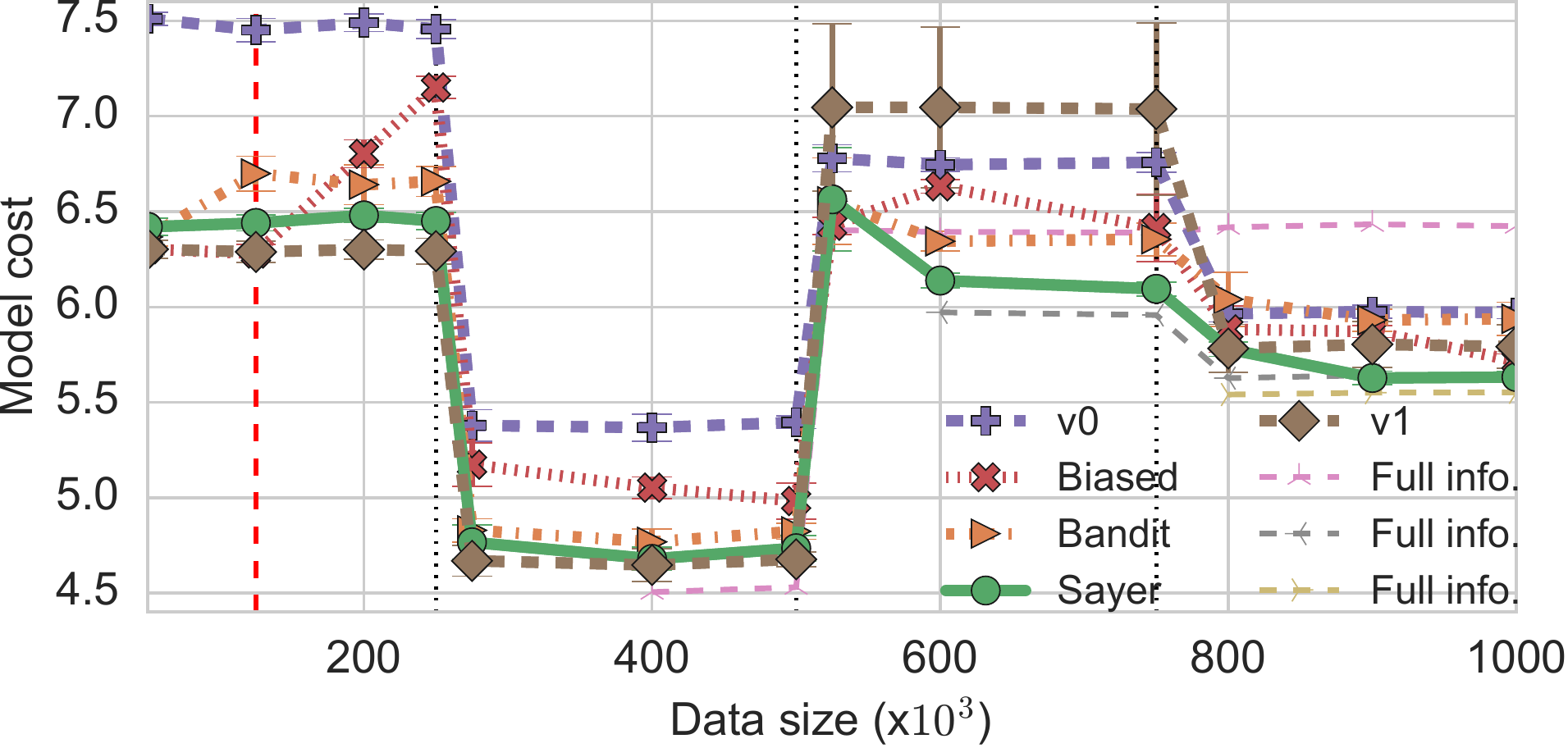}
    \caption{\footnotesize {\bf Evolution of performance}.}
    \label{fig:optim-full}
  \end{subfigure}%
  \hfill
  \begin{subfigure}{.43\textwidth}
    \centering
    \includegraphics[width=0.7\linewidth]{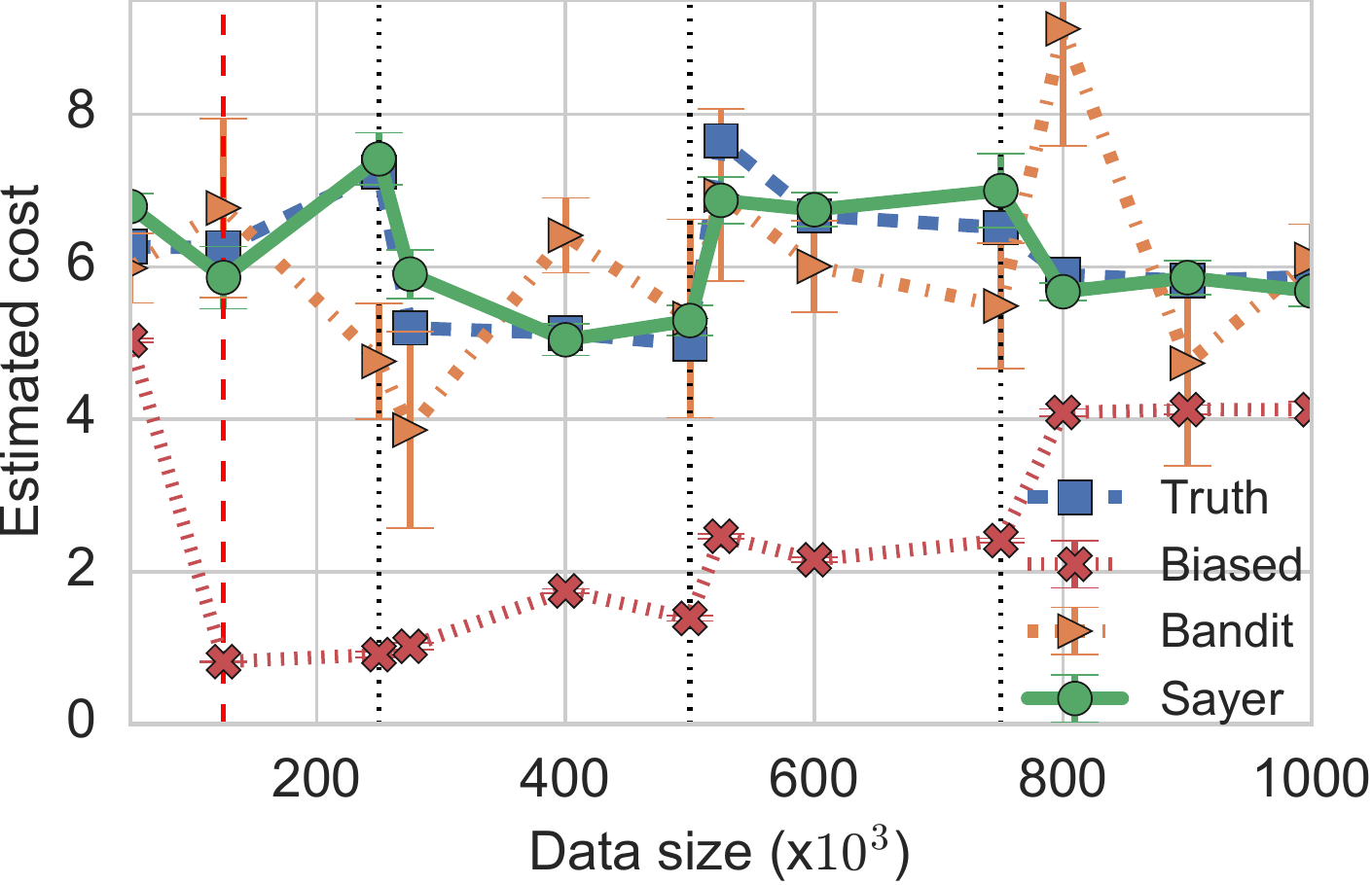}
    \caption{\footnotesize {\bf Counterfactual evaluation}.}
    \label{fig:counterfactual-full}
  \end{subfigure}
  \tightcaption{{\bf Behavior of periodic optimization and counterfactual evaluation.}
  The red vertical line shows when policies start being updated based on the data they collect.
  The black vertical lines show environment changes (the generator is trained on a different
  part of our data trace). 
  }
  \label{fig:fig}
\end{figure*}

\mypara{Counterfactual estimation accuracy}
Next, Figure~\ref{fig:counterfactual-barchart} evaluates the {\em accuracy} of various counterfactual estimators.
Counterfactual estimators can be used to evaluate any candidate policy, based on data collected when running a different policy.
To evaluate this capability, we use all our policies trained on \stwo, and evaluate their respective performance on \sthree (boxes in Figure~\ref{fig:counterfactual-barchart}) 
according to each counterfactual estimator (colored dots in Figure~\ref{fig:counterfactual-barchart}),
using data collected while running \oneshot.
Comparing these counterfactual results to the true (full information) cost, we can compute the bias (expected error) and variance of each counterfactual estimator. 
Accurate estimators can help operators distinguish good policies from bad ones without running the policies, based on data collected by the deployed policy.

\begin{itemize}

\item {\em Not accounting for missed feedback causes significant bias:} 
\direct, \naive, and \survival have mean estimates far away from the truth, even if they try to fill the gap of missing information (\direct and \survival).  
All three approaches also reverse the order of \sysname and other policies (not shown), resulting in a misleading assessment of their relative performance, 
and deploying a potentially worse policy if this assessment is acted on.
In contrast, \sysname and \ips use exploration and thus yield unbiased estimates when evaluated on multiple, varied policies.

\item {\em Implicit feedback reduces variance:} 
\ips, which uses exploration, also yields unbiased counterfactual estimates of a policy's cost, but compared to \sysname (\implicit estimator), the variance of
the estimate is much higher. 
This also implies poorer training performance, with the trained policy yielding similar results to the one shot model ($8.5\%$ improvement over \full) in Figure \ref{fig:optim-barchart}.

\end{itemize}

\subsubsection{Microbenchmarking using synthetic traces}
~

\noindent 
We also generate simulated traces inspired by the real dataset, in order to evaluate \sysname in the face of
non-stationarity, when periodically retraining policies on a trailing
window of data.
Our goal is not to model the real world exactly;
instead, our goal is to show how the different counterfactual analysis
techniques deal with non-stationarity. 

For realism, we learn the simulator's parameters from our production data as follows.
We draw from a Bernoulli distribution to determine if the machine is suffering from a temporary outage or a failure. 
For temporary outages, we draw $\tau$ from a Beta distribution with support in $[0, 10]$.
We create six scenarios by clustering recorded failures at different racks within a data center, and using the distribution of their recovery times in the full feedback data to learn the parameters for the Bernoulli and Beta distributions for each cluster.
The clusters correspond to different probabilities of recovery, and longer/shorter tails in recovery time.
We learn one such generator for each of the four splits in the trace described in \S\ref{s:eval-health-trace}. 
By switching from one generator to the next, we simulate a change in the environment.
Policies are initially trained on full feedback data, and then periodically retrained using a trailing window of the last $20k$ data points using \textit{only} the data they observe.
They are unaware of environment changes other than through the data.
Policies with exploration use an exploration rate of $10\%$.


Figure~\ref{fig:optim-full} shows the average downtime (cost) of the best performing baselines in Figure~\ref{fig:optim-barchart}: \ips, \direct, and \oneshot.
Environmental changes are shown by the three vertical dotted black lines.  The \omniscient lines show the performance of an omniscient policy based on full-feedback data started at the beginning of each period.
These are upper bounds on performance in their starting environment, but often degrade after environment changes.

Once again, deploying a policy without exploiting implicit feedback (\ie \ips) or not accounting for implicit feedback properly (\ie \naive) leads to poor performance that is often closer to the \full policy than to the
omniscient one.
The implicit feedback available to \sysname allows it to outperform the \ips-based policies by 3-18\%, depending on the environment. \sysname is also competitive with the \omniscient policy,
increasing downtime by only $3\%$, a relatively low cost.
Finally, the \oneshot policy performs competitively in the first and second
environments, but significantly underperforms in the third environment with a
downtime of $7$ minutes,
demonstrating the need for periodic retraining.

Figure~\ref{fig:counterfactual-full} shows the accuracy of counterfactual performance estimation on a single policy, using data generated by different deployed policies from the previous data window.
The bias of \naive again leads to an underestimate of up to $3\textrm{x}$. 
Compared to \ips, which is also unbiased, \sysname provides a significant reduction in variance (as shown by the bootstrap bars), allowing accurate estimation within smaller time windows.
This potentially reduces the time that stale models remain deployed in production environments.

\subsection{Application II: \scale}
\label{sec:scale}

\scale serves user requests to scale up a group of VMs by a given amount. 
When a user requests $k$ new VMs of a given type, to ensure the timely creation of these VMs, \scale over-allocates by creating $k + a$ VMs, and returns the first $k$ that are created. 
In this application, the goal is to trade off the resource cost of over-allocating VMs with the completion time $t$ of a requested group of VMs.
The over-allocation $a$ should be minimized to save resources, while being high enough to meet Service Level Objectives (SLOs) such as low median response time (MRT).
The default, conservative over-allocation policy (\full) deployed in \scale always chooses $a = 0.2k$ (or $a=k$ when $k$ is small, i.e. $\leq 4$),  and any over-allocation is capped to $0.2k$ (or $k$ when $k$ is small, i.e. $\leq 4$).

We use two possible cost metrics that capture this trade-off in \scale.
The first cost trades off meeting the MRT SLO with the over-allocation cost, formalized as:
\begin{align*}
  cost1 =  1\{t > MRT\} \cdot \min(t - MRT, max\_cost) + \gamma \cdot a
\end{align*}
where $MRT$ is the SLO objective (we use $80$ sec), $t$ is the creation completion time of the group of VMs, $max\_cost$ bounds the cost for stability (we use $100s$), and $\gamma$ is a factor to put $t$ and $a$ on the same scale (we use $13$).
Cost1 penalizes actions that miss the SLO ($t > MRT$) linearly up to $max\_cost$, while each over-allocated VM ``costs'' $\gamma$.
This cost enables implicit feedback in two forms. First, observing an over-allocation of $a$ VMs and
their completion times gives feedback for all over-allocations $a' \leq a$, since $t$ is
known for every $a'$ (but not for any $a'' \geq a$). Second, full feedback for every $a$ is given if
$t \leq MRT$, since the first term of the cost that includes $t$ is removed by the indicator
function.

We also use an alternative cost function with less implicit feedback:
\begin{align*}
  cost2  = 
  \min\Big(max\_cost, 0.99 \cdot \max(0, t-MRT) \\+ 0.01 \cdot \max(0, MRT-t) + \gamma \cdot a\Big).
\end{align*}
Here, both missing the $MRT$ and meeting the $MRT$ by too much are penalized, though with a smaller coefficient for meeting the $MRT$.
This cost receives less implicit feedback, because the information of larger actions is not revealed even when $t \leq MRT$, since we do not observe completion times for VMs we never created.
The only way to get full feedback is to choose the highest possible action.

To enable experimentation with different over-allocation policies, we build a prototype which mimics the functionality of \scale, using the public interface of \azure.
The prototype receives requests for $k$ VMs of a given type, decides on an over-allocation number ($a$), and issues $k+a$ VM creations to \azure.
It returns a completion time that is the $k^{th}$ smallest VM creation time to the user.
Unlike the production system, however, it also waits for all $k+a$ requests to complete and logs each completion time before deleting the additional VMs.
We use our prototype to replay an \scale production trace spanning 4 weeks, and containing 43821 requests made by a small subset of users to a North-American datacenter.
Requests are capped at $100$ new VMs, but most requests are small ($1$ to $8$ VMs).
The trace logs the request's timestamp, type of VM, and the over-allocation determined by the currently deployed policy.

\begin{figure}[t]
  \centering
  \begin{subfigure}{0.99\linewidth}
    \centering
    \includegraphics[width=0.85\linewidth]{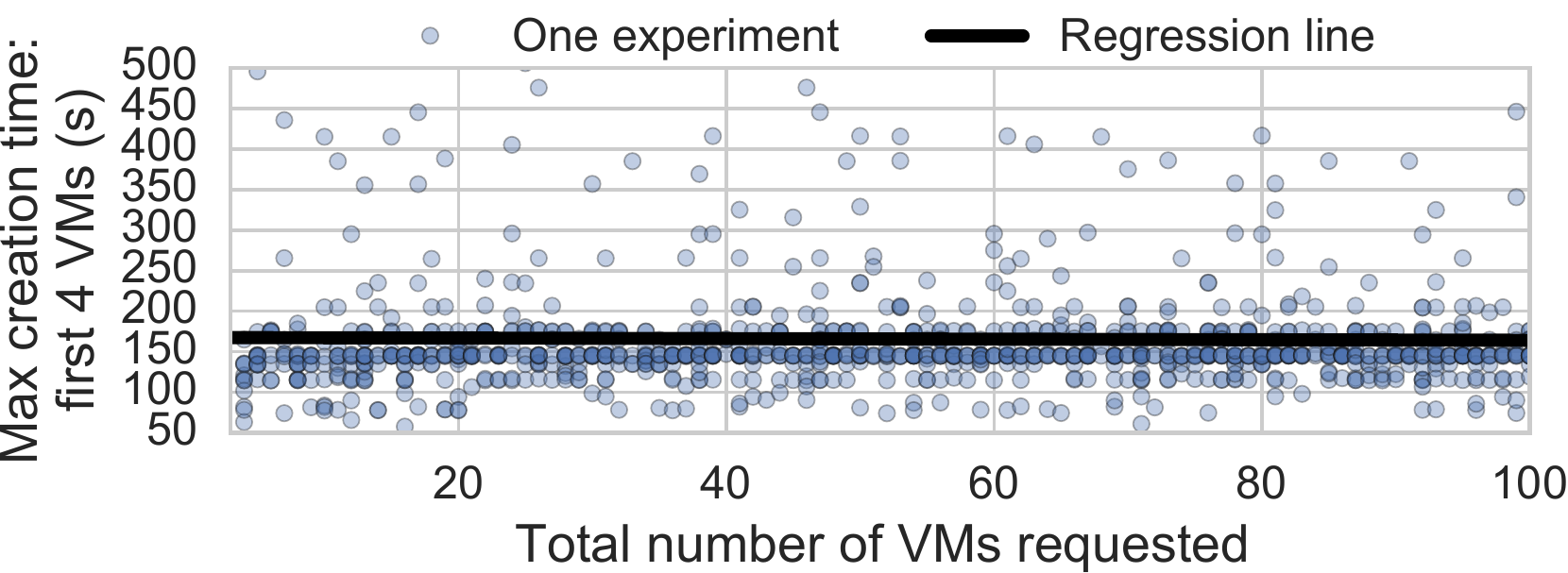}
  \end{subfigure}%
  \tightcaption{{\bf Assumption verified}: no significant influence of number of requested VMs on the completion times. 
  }
  \label{fig:causal-assumption-vmss}
\end{figure}

\mypara{Observed potential outcomes assumption}
Before we dive into the results, we first validate \sysname's observed potential outcomes assumption.
In the context of \scale, implicit feedback could be unreliable
if the system batches requests before starting allocations, and uses an allocation logic that depends on batch size. We designed a randomized experiment to empirically verify that the completion time of $k$ VMs out of the first $k+a'$ is the same whether we request $k+a'$ or $k+a, a>a'$. We sent $1000$ requests each with a randomly assigned total number of VMs ($k+a \in [1, 100]$, the maximum in our trace) and VM type, and analyzed the impact of the total requested number on the completion time of $k$ VMs out of $k+a'$. 
Figure~\ref{fig:causal-assumption-vmss} shows a representative example, for $k=4,a'=0$. For each value of $k$ and $a'$, we run a statistical test using a regression, and find no significant influence of total number of requests on the completion times: the slope of the regression is close to $0$, and the p-value is high, meaning that the data is compatible with our assumption.

\subsubsection{Trace-driven simulation}
~

\noindent We start by collecting a full feedback trace using our prototype, and split the resulting data into a training and testing set, each containing two weeks worth of data. We perform both policy optimization and counterfactual evaluation to test \sysname.
The key observations are as follows, which largely corroborate the takeaways from \health.

\begin{figure}[!t]
  \centering
  \begin{subfigure}{0.99\linewidth}
    \centering
    \includegraphics[width=0.85\linewidth]{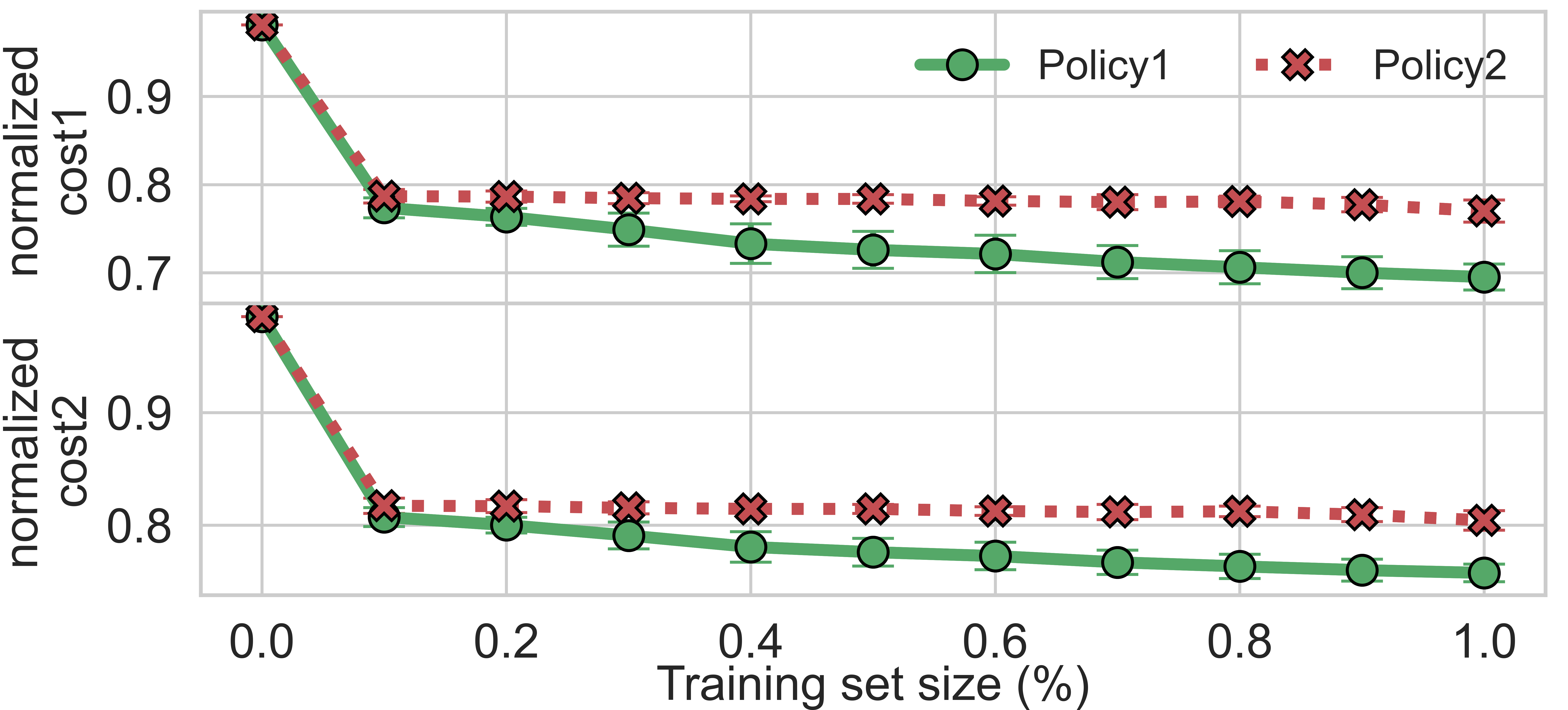}
  \end{subfigure}%
  \tightcaption{\textbf{Benefit of Implicit Feedback:}  Policy1 trained on cost1 (more implicit feedback) performs better than Policy2 trained on cost2 (less implicit feedback), when evaluated on either cost metric.
  }
  \label{fig:cost1-cost2-vmss}
\end{figure}

\mypara{Impact of implicit feedback}
To showcase the different values of implicit feedback, 
we use \sysname to train two policies, one to optimize cost1 (providing more implicit feedback) and the other for cost2 (providing less implicit feedback), respectively.
We train each policy using the log generated from a common \naive policy, with $10\%$ of exploration, on increasing amounts of training data from the first split, and evaluate their performance with full feedback on the second split.
Figure \ref{fig:cost1-cost2-vmss} shows that
Policy1, by leveraging implicit feedback, improves faster and performs better than Policy2 over the same amount of training data, showing 8.4\% and 6.3\% better performance in terms of cost1 and cost2 when all the training data is used.

\begin{figure}[!t]
  \centering
  \begin{subfigure}{0.99\linewidth}
    \centering
    \includegraphics[width=0.99\linewidth]{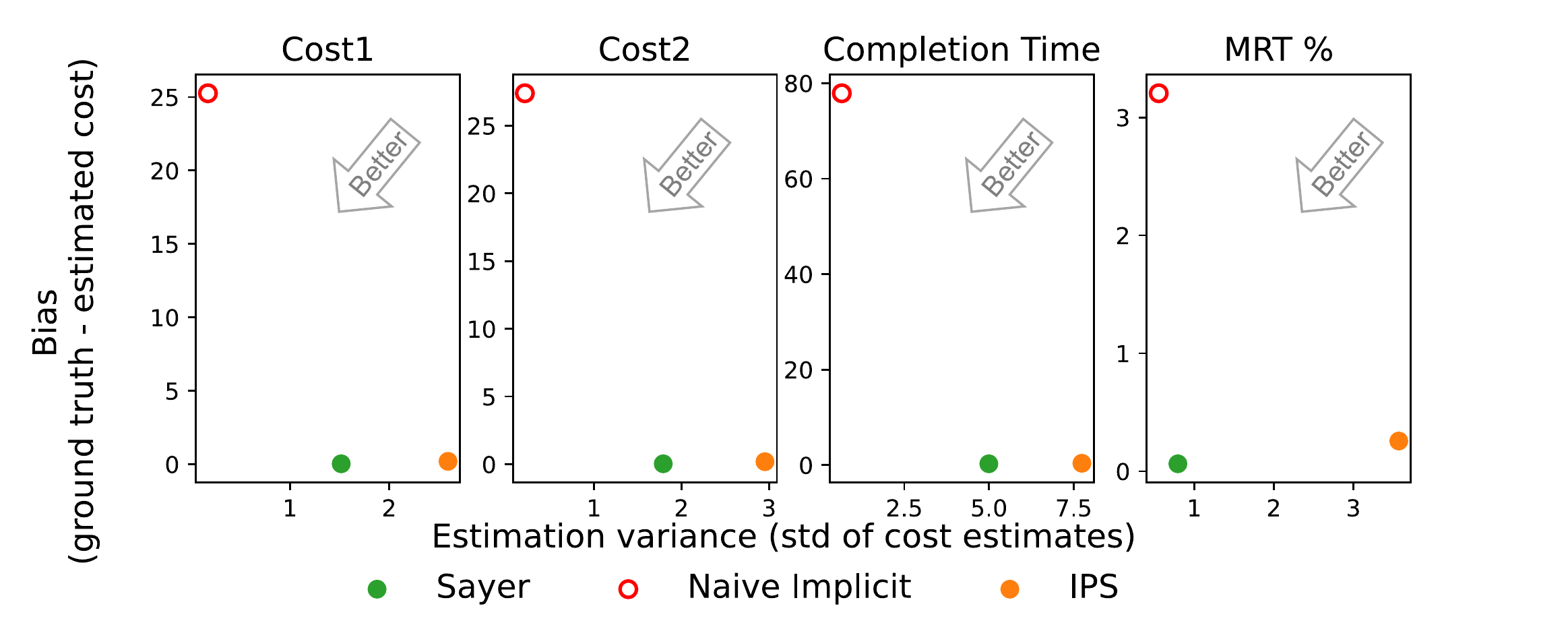}
  \end{subfigure}%
  \tightcaption{\textbf{Counterfactual Evaluation of Policy2} (cost1, cost2, completion time, MRT meet rate) shows that it can be used to estimate different metrics than the one optimized by the policy.} 
  \label{fig:policy2-eval-vmss}
\end{figure}

\mypara{Benefits of implicit exploration and feedback}
Figure~\ref{fig:policy2-eval-vmss} shows the counterfactual analysis error when using different approaches to evaluate Policy2's performance.
Each colored dot represents a different counterfactual estimator, but this time each box shows results for a different metric to evaluate (cost1, cost2, average VM creation time, and MRT meet rate).
This shows that counterfactual evaluation can be used to evaluate multiple practically useful metrics, and not just the cost optimized by the policy.
We can see that compared to \naive (using implicit feedback in a biased way) or \ips (unbiased but without implicit feedback), \sysname's estimates are unbiased and have a standard deviation only half that of the \ips estimator ($10$ sec vs $20$ sec), showing the value of both implicit feedback (compared to \ips) and exploration (compared to \naive).

\begin{figure}[!t]
  \centering
  \begin{subfigure}{0.99\linewidth}
    \centering
    \includegraphics[width=0.99\linewidth]{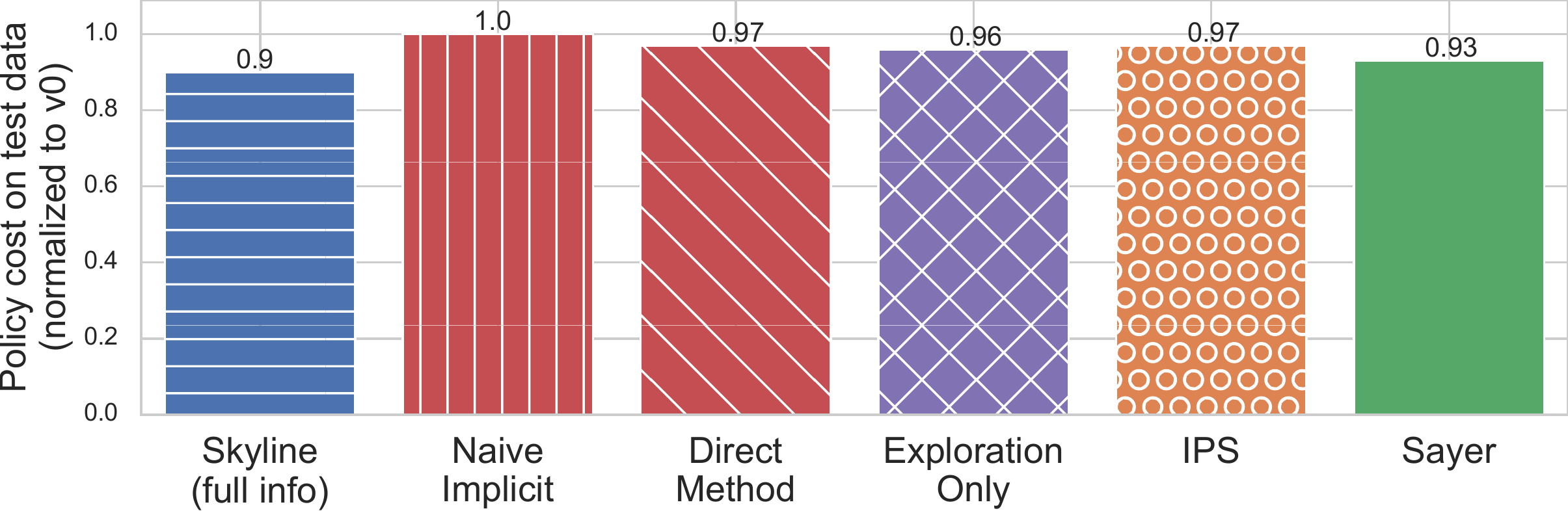}
  \end{subfigure}%
  \tightcaption{\textbf{Sayer vs Baselines (data-driven simulation in \scale).} 
  }
  \label{fig:baselines-vmss}
\end{figure}

\mypara{Benefits of \implicit estimator}
Finally, Figure~\ref{fig:baselines-vmss} shows results for policy training. \sysname outperforms policies trained by \naive, \direct, and \ips, by 7.0\%, 4.1\%, and 4.1\% respectively.
To emphasize the value of our \implicit estimator, we also add a policy trained exclusively on exploration data (Exploration Only in Figure~\ref{fig:baselines-vmss}). 
Such an approach is unbiased, but it cannot use datapoints in which the data collection policy did not explore. This data reduction yields a policy that is 3.1\% worse than \sysname, showing the value of our \implicit estimator leveraging every datapoint.

\subsubsection{Online evaluation using live deployment} 
~

\noindent Finally, we show the end-to-end performance of \sysname in a
{\em live deployment} of our prototype, compared to a \naive-trained policy,
and the default policy (\full).
Such an evaluation is not as simple as just deploying each policy in turn in the
prototype, because the strong temporal patterns in creation times within \azure
prevent comparisons between different time periods.
Consequently, we add support for simultaneously deploying policies and randomly assigning 
each request to one of them, in an A/B/C test.
This online testing framework allows \sysname and other policies to be deployed
in the same environment and compared on live traffic at the same time.  Both
policies are initially trained using the same one week of full-feedback data.
During the experiment, we replay our workload trace.
Every 24 hours, each policy is retrained on a 1-week trailing window of data.

\begin{figure}[!t]
  \centering
  \begin{subfigure}{0.99\linewidth}
    \centering
    \includegraphics[width=0.9\linewidth]{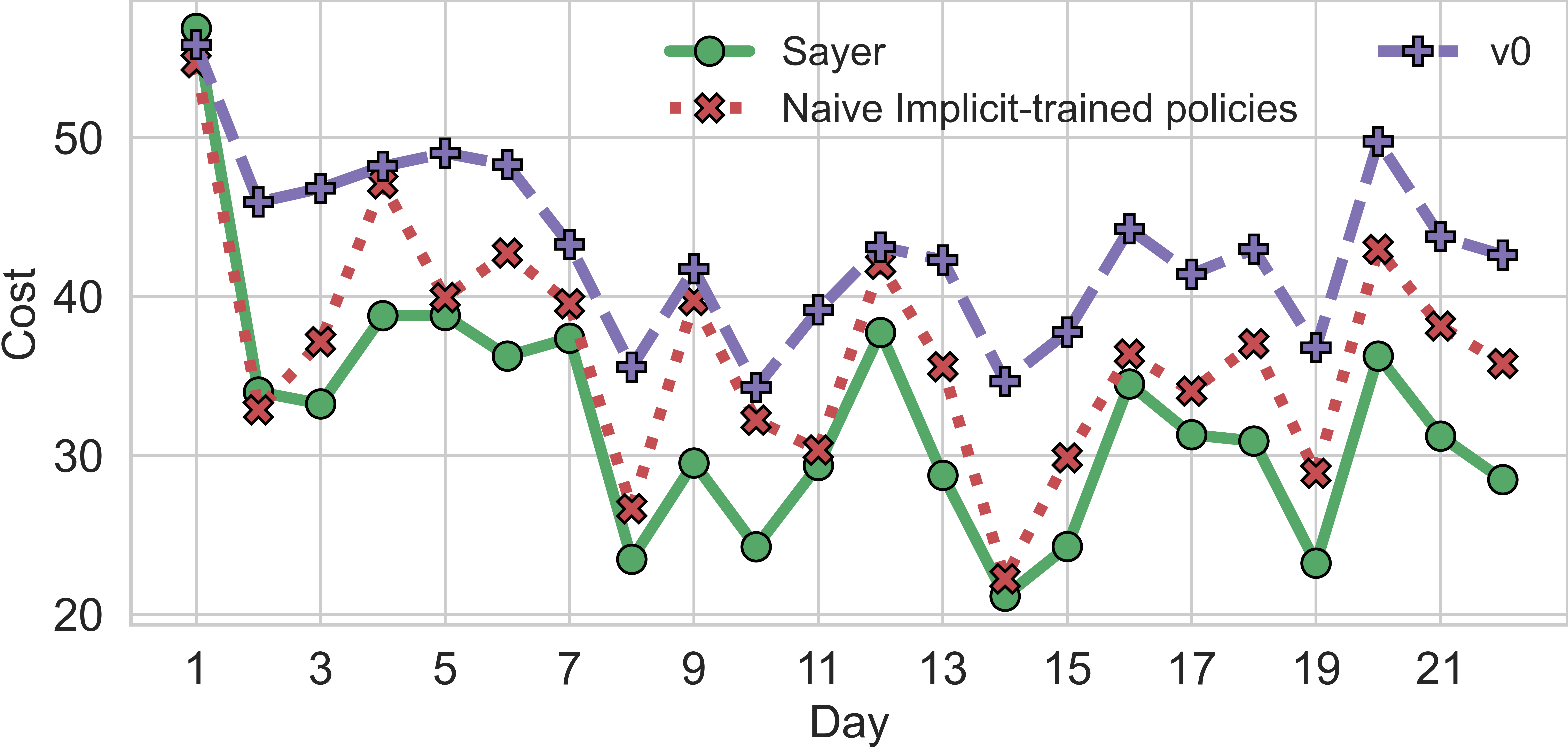}
  \end{subfigure}%
  \tightcaption{\textbf{Live deployment (A/B/C tests)}. Randomly assigning each request to \full, \naive, or \sysname. \sysname consistently has the lowest cost. 
  }
  \label{fig:online-vmss}
\end{figure}

Figure~\ref{fig:online-vmss} shows the performance of all three policies over
time.
At the beginning of the experiment, the \naive policy and \sysname
are both trained on full-feedback data and perform equally well.
Over time, the \naive-trained policy is retrained on biased data and performs
unevenly.
\sysname, on the other hand, consistently outperforms both the \full and the \naive-trained policy.

\ignore{
\section{Deployment concerns}

\sidn{Given our refocusing, it might be fine to move this to the discussion section at 
the end of the paper, replacing our current ``conclusion''.}
\sidn{I think people need some explanation for why we haven't deployed Sayer in the actual
production systems.  What other management concerns arise? Sometimes the cost of exploration is an issue;
sometimes its the fact that you are using certain kinds of ML policies to begin with, as
these models may be new for the team; sometimes its an education barrier as well that
needs to be worked through. In general we are asking product teams to accept a lot of new
things, and simply packaging it up in a framework like Sayer isn't going to give us
clearance to go forward and deploy. Consider a team that uses a heuristic policy; maybe
the first step for them is to include exploration and use that data to train whatever
policy they are using}
}


\section{Related work}
\label{sec:related}

\sysname attempts to bridge
the gap between recent theoretical work on counterfactual evaluation and
training in machine learning, and the challenge of evaluating and improving policies
in systems. We focus on the most related work from both sides, and point the reader to \S\ref{subsec:counterfactual}
for a broader overview.

\mypara{Counterfactual evaluation}
\sysname builds on Inverse Propensity Scoring (IPS) estimators, a classical technique
from the 1950s \cite{horvitz1952generalization,rotnitzky1995semiparametric}.
Recent work extends IPS to contextual bandits \cite{learning-from-ips} 
and incorporates inherent structure for better data efficiency~\cite{chu2011contextual,Langford-scavenge}.
These techniques have generally been applied to
advertising and news articles~\cite{bottou2013counterfactual,li2010contextual,Counterfactual-turorial-sigir16}.
\sysname adapts these techniques to design a methodology for
a broad class of system policies, and integrates it into their lifecycle.

\sysname leverages implicit feedback to boost the efficiency of counterfactual estimation.
Implicit feedback has been studied using feedback graphs~\cite{mannor-feedback,alon-feedback}, 
but these works assume a fixed feedback graph and use it to optimize one policy online.
In contrast, \sysname can counterfactually evaluate many policies, even when the feedback graph
is outcome-dependent. 

\sysname's implicit feedback can be viewed as a variance reduction technique over
IPS. Similarly, Doubly Robust (DR) estimators can also be used to reduce the variance 
of IPS, by combining a model-based predictor to 
``fill the gaps'' when information is missing~\cite{dudik2014doubly}. 
DR estimators are orthogonal to our contribution and can be applied to both \sysname and IPS. 

\mypara{Data-driven modeling in systems}
While most work in systems is evaluated on real testbeds or deployments, trace-driven
evaluation is often used to evaluate new policies at scale (e.g., in
server/path selection~\cite{liu2016efficiently,jiang2016via}, video bitrate
adaptation~\cite{mao2017neural,jiang2016cfa,yin2015control}, and MapReduce
scheduling~\cite{kumar2016hold}).
Data-driven models/simulators were developed for ``what-if'' analysis in specific
systems settings (e.g., web service~\cite{jiang2016webperf}, CDN server
selection~\cite{tariq2008answering}, end-to-end adaptation performance~\cite{sruthi2020pitfalls}, and streaming video QoE
modeling~\cite{krishnan2013video}). 
Similar data-driven performance modeling is also used to predict the best configuration for a workload based on a few samples~\cite{zhu2017bestconfig}.
However, it is inherently difficult for 
these analyses to faithfully capture all relevant details (including confounding factors~\cite{sruthi2020pitfalls}) of a large-scale 
system~\cite{floyd2001difficulties}, to simulate or build an analytical model that precisely predicts performance of any unobserved actions~\cite{fu2021use}.  In contrast, \sysname focuses on a class
of decisions that exhibit independence properties, and uses tools from statistics/machine learning to enable unbiased evaluation without the need for modeling.
Recently,~\cite{lecuyer2017harvesting,bartulovic2017biases} suggested the potential
of such an approach, but fall short of addressing any systems challenges or
developing any usable methodology.

\mypara{RL and Online Learning in systems}
A closely related body of work uses (deep) reinforcement learning
(e.g.,~\cite{tesauro2007reinforcement,mao2017neural,erickson2010effective,alipourfard2017cherrypick,van2017automatic,li2019qtune,zhang2019end})
or online learning
(e.g.,~\cite{dong2015pcc, dong2018pcc})
in systems optimization.
It optimizes a {\em single policy} online by
continuously interacting with the environment.
Typically, the data collected by such a policy yields partial
feedback that can only be used to evaluate policies
that are similar to it. 
\sysname instead focuses on a class of techniques 
that enable unbiased counterfactual evaluation of {\em any policy}.
Although our evaluation focuses on iterative model updates (which are common in production systems), 
we note that \sysname can also update a policy online as new data arrives.

\section{Discussion}
\label{sec:discussion}

We applied \sysname's counterfactual evaluation and training methodology to two systems:
\health and \scale. These examples illustrate the value of counterfactual evaluation in
systems, as well as the prevalence of implicit feedback in systems that make threshold
decisions. They also illustrate the manual effort required to apply \sysname:
specifically, \sysname relies on the system designer to define an event $E$ that captures
the available implicit feedback, and compute its probability $P(E)$. Though nontrivial, defining
this event follows naturally from reasoning about when cost feedback is known for the 
candidate policy's action, based on the data collected by the deployed policy. 
For example in \health, the only case when feedback is not available is when the deployed
policy's action causes a timeout and the candidate 
policy chooses to wait even longer; $E$ is thus defined as the opposite of this event.
After deriving $E$ for \health, it was relatively straightforward to do the same for \scale.
Thus, in our experience, the manual effort required for a new application is reasonable. 

As discussed in \S\ref{subsec:design:overview}, \sysname applies to system decisions 
that satisfy certain independence properties (\eg the same ones required by contextual bandits and \ips).  
As such, it does not apply to system policies that maintain long-term state, or
whose decisions interact in complex ways. 
Defining appropriate events and costs for individual actions in such settings is a 
challenging problem for reinforcement learning.
It is an interesting open question if the ideas from \sysname, and in particular
our techniques for leveraging implicit feedback, can be extended to more general
RL to support these settings.

\sysname focuses on one-dimensional (single-parameter) decisions due to two limitations
in our approach. The first is our reliance on existing RL techniques, which do not cope 
well with large or complex action spaces, mainly because they are unable to explore these 
spaces efficiently. A multi-dimensional action space grows exponentially in the number of
dimensions: \eg even a 2-parameter decision $(x \in \mathcal{X},y \in \mathcal{Y})$ where $|\mathcal{X}|=|\mathcal{Y}|=N$ has
an action space of size $N^2$. The
second limitation is that it is unclear if implicit feedback can be obtained in
multi-dimensional decisions. For example, if we take the decision $(x,y)$, does that mean that we
receive feedback for all actions $(\le x, \le y)$? Clearly this depends on the
relationship between $x$ and $y$, which may be complex.
Extending Sayer to multi-dimensional action spaces is an interesting direction for future
work.

\bibliographystyle{ACM-Reference-Format}
\bibliography{reference}

\end{document}